\documentclass[final,12pt]{clear2024} 

\title[]{Causal State Distillation for Explainable Reinforcement Learning}
\usepackage{times}

\usepackage{enumitem}
\usepackage{wrapfig}

\SetKwComment{Comment}{/* }{ */}

\usepackage{booktabs, makecell, multirow}
\usepackage{caption}
\usepackage{pifont}
\usepackage[section]{placeins}

\newcommand{\NA}{---}
\newcommand{\Mod}[1]{\ (\mathrm{mod}\ #1)}

\newcommand{\cmark}{\ding{51}}%
\newcommand{\xmark}{\ding{55}}%

\DeclareMathOperator*{\argmax}{arg\,max}

\clearauthor{
 \Name{Wenhao Lu} \Email{wenhao.lu@uni-hamburg.de}\\
 \Name{Xufeng Zhao} \Email{xufeng.zhao@uni-hamburg.de}\\
 \Name{Thilo Fryen} \Email{thilo.fryen@uni-hamburg.de}\\
 \Name{Jae Hee Lee} \Email{jae.hee.lee@uni-hamburg.de}\\
 \Name{Mengdi Li} \Email{mengdi.li@studium.uni-hamburg.de}\\
 \Name{Sven Magg} \Email{sven.magg@uni-hamburg.de}\\
 \Name{Stefan Wermter} \Email{stefan.wermter@uni-hamburg.de}\\
 \addr University of Hamburg
}

\begin{document}

\maketitle

\begin{abstract}
    Reinforcement learning (RL) is a powerful technique for training intelligent agents, but understanding why these agents make specific decisions can be quite challenging. This lack of transparency in RL models has been a long-standing problem, making it difficult for users to grasp the reasons behind an agent's behaviour. Various approaches have been explored to address this problem, with one promising avenue being reward decomposition (RD). RD is appealing as it sidesteps some of the concerns associated with other methods that attempt to rationalize an agent's behaviour in a post-hoc manner. RD works by exposing various facets of the rewards that contribute to the agent's objectives during training. However, RD alone has limitations as it primarily offers insights based on sub-rewards and does not delve into the intricate cause-and-effect relationships that occur within an RL agent's neural model. In this paper, we present an extension of RD that goes beyond sub-rewards to provide more informative explanations. Our approach is centred on a causal learning framework that leverages information-theoretic measures for explanation objectives that encourage three crucial properties of causal factors: \emph{causal sufficiency}, \emph{sparseness}, and \emph{orthogonality}. These properties help us distill the cause-and-effect relationships between the agent's states and actions or rewards, allowing for a deeper understanding of its decision-making processes. Our framework is designed to generate local explanations and can be applied to a wide range of RL tasks with multiple reward channels. Through a series of experiments, we demonstrate that our approach offers more meaningful and insightful explanations for the agent's action selections.
\end{abstract}



\section{Introduction}

Many efforts have been made to adapt \textit{post-hoc} saliency approaches from the field of explainable machine learning~\citep{DBLP:journals/corr/SelvarajuDVCPB16, DBLP:journals/corr/RibeiroSG16, shrikumar2017learning, sundararajan2017axiomatic} to understand the behaviour of reinforcement learning (RL) agents. These approaches usually aim to provide visual explanations by highlighting salient state features that influence an agent's action choices \citep{greydanus2018visualizing, DBLP:journals/corr/abs-1809-06061}.

However, we identify two key issues in applying these approaches to RL. First, there is a general concern about using saliency maps to explain RL agent behaviour as post-hoc explanations are not grounded in the agent's learning process~\citep{milani2022survey}.
The work by~\cite{atrey2020exploratory} emphasizes that saliency might convey misleading, non-causal interpretations of agent actions. 
For example, in Breakout, the saliency pattern and intensity around a tunnel vanish when a reflection intervention is applied to bricks near the tunnel, refuting the hypothesis that agents learn to aim at tunnels~\citep{atrey2020exploratory}.


\begin{figure}
    \centering%
    \includegraphics[width=0.75\textwidth]{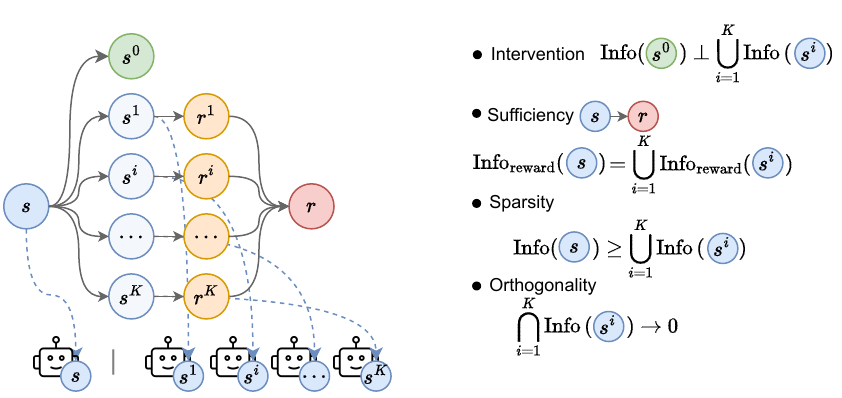}
    \caption{The disentanglement of state representations and resulting sub-agents when uncovering the cause-effect relationships with \textit{causal state distillation} (action omitted for brevity). Here, $s^0$ denotes the distilled non-causal components of state $s$, while $s^i$ captures the causal elements, each linked to a distinct reward aspect $r^i$. Sub-agents focus on a singular causal component for policy learning. The distillation process, consisting of multiple learning steps, is elaborated in Sec.~\ref{sec:learning framework}.}
    \label{fig:rdflow}
\end{figure}

Second, saliency-based approaches often overlook RL-specific aspects, limiting their effectiveness in generating meaningful saliency maps. They are developed for supervised tasks, which typically address non-temporal reasoning and are focused on model behaviours concerning specific objectives, such as classifying an image into a specific category. Explanations for RL agents must go beyond this and provide additional insights into the agent's interaction data, encompassing the rewards it has received, the states it has transitioned between, and the diverse goals it strives to achieve. This contextual information, which exists during the learning process, is vital for refining our understanding of the agent's decision-making. Unfortunately, saliency maps fall short in this regard, as their generation does not rely on any interaction data.

In this research paper, we thus take a new route and investigate a way to allow RL agents to \textit{intrinsically} attend to \textit{causal} but \textit{distinguishable} state components, predictive of the agent's action and reward obtained during its learning. An appropriate candidate we consider here is \emph{Reward Decomposition} (RD)~\citep{Juozapaitis19ExplainableReinforcement, septon2023integrating, lu2023closer} which discerns the contribution of each sub-reward to the agent's decision-making. However, RD has its limitations, as it does not unveil which specific state components are being utilized or attended to by each decision-making policy induced by various sub-rewards. Our primary focus is on RL tasks where there are multiple reward channels (i.e., sub-rewards) sourcing from different environmental factors, for example, both bonking the gopher or filling holes contribute to the achievement of the goal in the Gopher game~\citep{DBLP:journals/corr/abs-1207-4708}.

To ensure that we attain various sorts of attention from the agent that faithfully explains its decision-making process, a powerful approach is to use the language of causality. In this paper, we introduce a structural causal model~\citep{10.5555/1642718} to formalize the problem of how different state components contribute to diverse reward aspects or, as a consequence, Q-values (see Fig.~\ref{fig:rdflow} for an overall visualization). Concretely, we aim to separate the latent factors (or \emph{state components}) that are causally relevant to the agent's decision-making from those that are not. Besides, we introduce three desired properties of causal factors, i.e., \textit{sufficiency}, \textit{sparsity} and \textit{orthogonality}, to constrain the information flow during the learning process. An inherent advantage of our explanatory framework is that the learned causal factors can serve as a rich vocabulary for explicating an agent's action. These causal factors improve over saliency maps in both expressiveness and diversity. Each latent factor, in isolation, unravels intricate patterns (events) in the agent's interactions. Moreover, this ensemble of diverse factors offers a multifaceted perspective on the agent's attention to each of them, thereby unveiling the rationale behind its actions.

Our contributions can be summarized as follows:
\begin{itemize}
    \item We investigate RL explanations from a causal perspective and propose a novel framework for generating explanations in the form of causal factors, driven by three essential desiderata.
    \item We present two paradigms (R-Mask and Q-Mask) of distilling causal factors, in which the factorization is ensured by imposing causal sufficiency of reward and Q-value respectively.
    \item We establish reasonable evaluation metrics to quantify the explanatory quality.
    \item We conduct an analysis of this framework in a toy task for intuitive understanding and an extended evaluation applied to explaining agents involved in complex visual tasks.
\end{itemize}

\section{Related Work}


In line with the taxonomy of eXplainable Artificial Intelligence (XAI) approaches, XRL approaches can be naturally categorized into two scopes: \textit{local} and \textit{global}. Local approaches refer to explaining a single decision for a single situation. In contrast, global approaches aim to explain the long-term behaviour of a learned RL model (i.e., on policy or trajectory level)~\citep{milani2022survey, qing2022survey}. Our explanation framework globally learns to discover which state components (latent factors) are beneficial for local explanations.

\paragraph{Local Feature Importance.} Most local explanation techniques for RL extend from those in XAI, explaining the prediction for a specific data instance~\citep{DBLP:journals/corr/SelvarajuDVCPB16, DBLP:journals/corr/RibeiroSG16, shrikumar2017learning, sundararajan2017axiomatic}. Those local explanations provide action-oriented explanations for RL agents' behaviour through post-hoc \textit{rationalization}. Post-hoc interpretability refers to generating action explanations for a non-interpretable RL model, by the forms of saliency maps~\citep{greydanus2018visualizing, DBLP:journals/corr/abs-1809-06061, DBLP:journals/corr/abs-1912-12191}. The work of~\cite{greydanus2018visualizing} derives saliency maps by observing the changes in the policy after adding Gaussian blur to different parts of input images. However, the saliency map can highlight regions of the input that are not relevant to the action taken by the agent. Complementary saliency work by~\citet{DBLP:journals/corr/abs-1912-12191} mitigates this issue.
Nevertheless, the saliency map used in practice as evidence of explanations for RL agents might be highly subjective and not falsifiable~\citep{atrey2020exploratory}. That is, \textit{ad hoc} claims to the agent's behaviour are proposed after the presented saliency is interpreted.



\paragraph{MDP-aware Explanation.} Another important category of explanations is those which expose the impact of parts of the MDP (e.g., reward $\mathcal{R}$ and dynamics model $\mathcal{P}$)~\citep{puterman2014markov} on the agent's behaviour. Those techniques generally require additional information for training. For example, the line of work in \textit{reward decomposition}~\citep{Juozapaitis19ExplainableReinforcement, septon2023integrating, lu2023closer} needs to know the existing reward structure prior to the agent's learning. The resulting explanation artefacts clarify the contribution of each reward component to the agent's decision (i.e., Q-values). However, despite their potential, these explanations rely on scalar Q-values and do not disclose which state aspect impacts the estimation of diverse Q-values, limiting their actionable value.


\paragraph{Causality in Explanations.} The language of cause and effect has gathered increasing attention in generating explanations~\citep{10.1145/3400051.3400058}. An earlier work for causal explainability is~\cite{DBLP:journals/corr/abs-1905-10958}, which explains ``why'' and ``why not'' questions with a learned causal model. However, it relies on known abstract state variables for explanation generation, restricting it to discrete setups. On the contrary, our approach extends to continuous settings, accommodating learned factors, associated with various reward facets, for explanations. The work of~\cite{DBLP:journals/corr/abs-2007-13531} aims to achieve a parameterizable interpretation of the expert’s behaviour in the batch Inverse Reinforcement Learning (IRL) setting by employing a counterfactual-based reward function. However, this method is limited to linear reward functions based on data features. 
Recent work has quantified state and temporal importance to action selection by leveraging learned structural equations for a known causal structure~\citep{wang2023causal}, but its application is limited to unusual cases with abstract states. Unlike it, some aim to find explanatory input (e.g., graph or image data) for model prediction by measuring information flow (which can be seen as the causal counterpart of mutual information)~\citep{ay2008information, DBLP:journals/corr/abs-2006-13913, lin2022orphicx}, or by causal interventions~\citep{lv2022causality, 10.1145/3580305.3599240}. 
However, they provide merely post-hoc causal explanations within the realm of supervised settings. In contrast, our research delves into the realm of \textit{inherent} RL explanations, a more intricate problem, approached from a causal perspective. Though our proposed causal RL explanation framework draws upon similar notions of causality as found in non-RL post-hoc explanation works by~\cite{DBLP:journals/corr/abs-2006-13913, lin2022orphicx} for constructing explanations, we emphasize that ours is unique in that the framework can generate latent factor-based explanation associated to various reward facets, all the while coevolving with the agent's policy learning. 

Our explanation method can be categorized into causal RL~\citep{Zeng23SurveyCausal}, an emerging subfield of RL that harnesses the power of causal inference. Different from approaches~\cite{Zhu_Jiang_Liu_Yu_Zhang_2022, guo2022relational} which utilize causality for learning representation that benefits the generalizability and sample efficiency of RL agents, this work leverages causality to learn an intrinsically interpretable RL policy.



\section{Methodology}

Our goal is to locally explain an agent's action at a state from the causal view with a structural causal model (SCM)~\citep{10.5555/1642718}
that globally describes how \textit{factors} or components
of states $\langle \alpha, \beta \rangle$ \textit{causally} affect agent's actions and rewards it received. The effect is causal in the sense of changing the causal factors $\alpha$ produces changes in the agent's behaviour and the consequence, while non-causal factors $\beta$ should not.

To formalize the explanations, we need to define (i) a causal graph
that relates state factors, agent's actions ($a$), and its rewards ($r$); (ii) an approach to disentangling causal factors from non-causal ones; (iii) a metric to measure the causal influence of $\alpha$ on $a$ and $r$; and (iv) a learning framework that learns $\alpha$ while ensuring the success of the policy learning of corresponding RL tasks. Here, we focus on RL tasks with multiple reward channels which may be unknown during training.



\begin{figure}[ht] 
    \centering
    
    \subfigure[The SCM for RL explanations where $\alpha$ represents causal factors and $\beta$ for non-causal ones.]{%
    \includegraphics[width=0.28\textwidth]{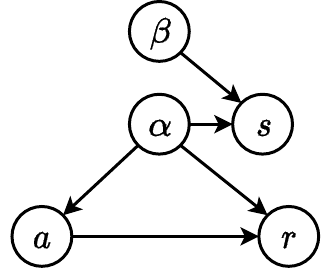}
    \label{fig:scm}
    }
    \hfill
    \subfigure[The extended SCM for RL explanations with an example illustration of factors and sub-rewards (right). For instance, both $\alpha^1$ and $\alpha^2$ determine sub-reward $r^2$. For brevity, the edges from all subsets of factors ($\bar{\alpha}^i$) to action $a$ are omitted (rightmost part).]{%
        \includegraphics[width=0.62\textwidth]{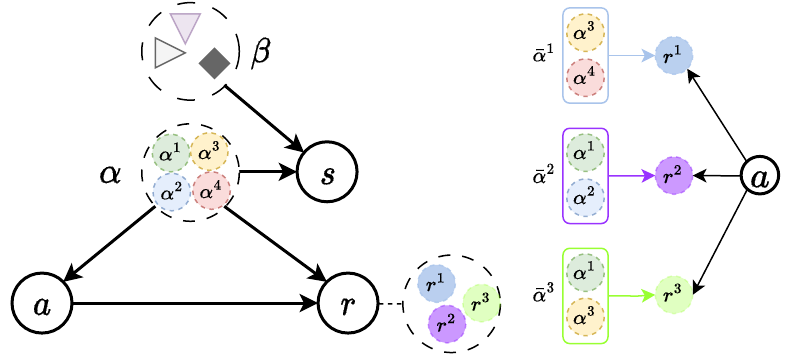}
        \label{fig:causal factors}
    }
    
    \caption{
      The causal graph for one-step RL explanations.
    }
  \label{fig:scm for causal explanations}
\end{figure}

    

\subsection{A causal view on explanations} \label{sec: a causal view}

Our explanations for agent's behaviours take the form of a set of causal factors. That is, by construction, the functional relationship defining the causal connection $\pi: \alpha \to a$ 
uses only the factors of a state $s$ that are causal. Based on this observation, we then adopt an SCM as depicted in Fig.~\ref{fig:scm} to describe the causal structure between $\alpha$, $a$, and $r$. In tandem with $\alpha$, non-causal factors $\beta$ contribute to representing states the agent observed but would not causally influence the agent's actions and rewards. Stated differently, any interventions on $\alpha$ and $\beta$ cause changes in $s$, but only interventions on $\alpha$ cause changes in $a$ and $r$. Besides, any alternations to $\beta$ would not have an impact on the causal factors $\alpha$ as well. Importantly, we do not \textit{assume} $\alpha$ is given a priori as~\cite{7546525, shrikumar2017learning} do, but we intentionally \textit{learn} to separate $\alpha$ from $\beta$. A formalization of the RL problem in SCM can be found in Appendix~\ref{sec:forma-rl-scm}.



Since causal factors are generally not observable~\citep{arjovsky2019invariant} and their extraction relies on the availability of specific supervision signals and interventions~\citep{DBLP:journals/corr/abs-2102-11107}, we seek to learn them in a way that each factor in $\alpha$ corresponds to a different aspect of the environmental state and a subset of causal factors has a sizeable causal influence on a reward component (sub-reward) $r^i$ and the action chosen $a$. To this end, we expand the SCM in Fig.~\ref{fig:scm} to explicitly illustrate the relationship among causal factors, action, and sub-rewards, as depicted in Fig.~\ref{fig:causal factors}. 



\subsection{Notions and desiderata for explanations}
\label{sec:noti-desid-expl}

\textbf{Notions.} We assume a factorization of $\alpha = \{ \alpha^1, \alpha^2, \dots, \alpha^N\}$ and the additivity of reward $r = \sum_{i=1}^K r^i$, where $N, K\in \mathbb{N}$. Notably, $N$ and $K$ may differ. We further denote $\bar{\alpha}^i$ as a subset of causal factors corresponding to a sub-reward $r^i$
and the actual values of sub-rewards may be unknown a priori. 
As for retrieving causal factors $\bar{\alpha}^i$, we extract them from the raw state $s_t$ or a learned representation of it, i.e., $\alpha = \psi(s_t)$ by using a neural network-based masker $m^i(\cdot)$, i.e., $\bar{\alpha}^i = m^i(s_t)* \psi(s_t) = m^i(s_t) * \alpha$.



To ground the learning of causal factors $\alpha$ functioning as described in Sec.~\ref{sec: a causal view},
we further highlight several desiderata for explanations that these learned factors are expected to fulfill.
In the next sections, we discuss how to approach these desiderata from the standpoint of information theory and by using do-operator $\textit{do}(\cdot)$~\citep{Pearl2016CausalII}.

\begin{itemize}
    \item 
    The causal factors $\alpha$ should be independent of non-causal factors $\beta$, i.e., $\alpha \perp \beta$. Thus, intervening on $\beta$ does not change $\alpha$ and the learned $\pi: \alpha \to a$ as well.
    \item 
    The causal factors $\alpha$ (or $\bar{\alpha}^i$) are desired to be causally sufficient for rewards $\alpha \to r$ (or sub-rewards $\bar{\alpha}^i \to r^i$) and action $\alpha \to a$, i.e., to contain all information required to predict $r$ (or $r^i$) and explain the causal dependency between $\alpha$ and $a$.
    \item 
    Given any two subsets of causal factors $\bar{\alpha}^i$, $\bar{\alpha}^j$ corresponding to sub-rewards $r^i$ and $r^j$ respectively, $\bar{\alpha}^i$ (or $\bar{\alpha}^j$) needs to contain less or no information about determining $r^j$ (or $r^i$). Besides, we expect $\bar{\alpha}^i$ (or $\bar{\alpha}^j$) to be minimally sufficient, i.e., containing the least amount of (sufficient) information for predicting $r^i$ (or $r^j$).
\end{itemize}




\subsection{The learning framework}\label{sec:learning framework}

Recall that the first criterion indeed amounts to performing the causal intervention~\citep{10.5555/1642718} on non-causal factors $\beta$, i.e., $P(\alpha|\textit{do}(\beta))$, 
the second requires a metric for the causal influence of $\alpha$ on $a$ and $r$ using the SCM in Fig.~\ref{fig:causal factors}, and the last needs a
measure of independence between any subsets over $\alpha$. Together, a learning framework is developed to unify these desiderata.

\subsubsection{Metric for causal intervention}

In general, causal and non-causal factors coexist in the agent's interaction with the environment. We aim to separate causal factors $\alpha$ from non-causal ones $\beta$ by causal intervention, ensuring that $\alpha$ remains invariant when $\beta$ undergoes interventions ($\textit{do}(\beta)$). Notably, non-causal factors may not always be directly observable but can be accessed through domain knowledge. For instance, in Atari games, the displayed scores on the scoreboard can be considered a non-causal factor. 
As~\citet{piotrowski1982demonstration} noted, the Fourier transformation preserves high-level semantics in the phase component while encoding low-level statistics in the amplitude component. Therefore, in line with~\citet{lv2022causality}, we intervene on $\beta$ by perturbing the amplitude component while maintaining the phase. Starting with the original state $s$ and a state $s^\prime$ devoid of non-causal factors, we perform the intervention, resulting in an intervened state $s^\textit{inter}$ (i.e., $s \backslash \beta$, where $\beta$ associated parts are removed). Details on the intervention procedure are available in Appendix~\ref{sec:computing-ca-suff}. Then, we optimize the encoder $\psi$ by maximizing the following correlation to maintain the invariance of $\alpha$ following the aforementioned intervention upon $\beta$:

\begin{equation}\label{eq:l0}
    \max \sum_i cos(\psi(s), \psi(s^\textit{inter})),
\end{equation}
where we leverage cosine similarity $cos(\cdot, \cdot)$ to measure the correlation between causal factors before and after intervening on $\beta$. 




\subsubsection{Metric for causal sufficiency}\label{sec:me-ca-suff}


\textbf{Causal sufficiency for reward.} A distilling masker ${m^i(\cdot)}$ is regarded as causally sufficient if the information transition to the reward is sufficient such that 
the causality between the (sub-)event trigger and its environmental feedback holds clearly, i.e., $\mathbb{E} \log \hat{p}(r^i|\bar{\alpha}^i_t)=\mathbb{E} \log p(r^i|s_t)$ and $\mathbb{E} \log \hat{p}(r|\bigcup_{i=1}^{K}\bar{\alpha}^i_t) = \mathbb{E} \log p(r|s_t)$. 
The sufficiency 
of $\bar{\alpha}^i$ to deduce $r^i$ can be achieved by maximizing their mutual information $\mathcal{I}(\bar{\alpha}^i; r^i)$
or fitting a reward model $\mathcal{R}_\theta$ such that $r^i = \mathcal{R}_\theta(\bar{\alpha}^i, a)$.
The total information regarding the environmental causality thus can be persisted via the regression 
$r = \sum_i^{K} r^i = \sum_i^K \mathcal{R}_\theta(\bar{\alpha}^i, a)$, 
i.e., by minimizing the $L_2$-norm fidelity loss
\begin{equation}\label{eq:l1}
   \min \mathbb{E} \Vert \sum_i R_\theta(\bar{\alpha}^i, a) - r \Vert_2,
\end{equation}
towards reward information persistence (omitted when raw sub-rewards are given in advance).


\textbf{Causal sufficiency for action.} 
Though, by disentangling state representation with the above objective we can obtain causal factors that are sufficient in terms of determining sub-reward $r^i$, it is equally crucial to get the impact of causal factors timely involved in action selection, i.e., whether the distilled factors are sufficient or even beneficial for learning an optimal policy. The joint learning process of decomposing state and fitting a policy may fall into an unstable or even vicious loop --- insufficient factors exert challenges to policy learning, while non-informative trajectories unrolled by an under-optimized policy, in turn, hinder the causality distillation~\citep{li2023irrl}. We thus report the findings of (masked) Q-learning with causal factors under the setting that sub-rewards are \textit{known} from the environment, leaving the more challenging one, where the reward decomposition has to be jointly learned, for future work.

To assess the impact of causality distillation on Q-agent learning in RD, we contrast two controlled Q-learning variants with and without access to the full state. That is, the Q-agent consumes and updates according to the sub-state (that is sufficient and concise to reveal the $i$-th causal aspect of the state): $Q^i(\bar{\alpha}^i_t, a_t) \leftarrow (1-\alpha ) Q^i(\bar{\alpha}^i_t,a_t)+ \alpha [r^i_t + \gamma Q^i(\bar{\alpha}^i_{t+1},a_t^\ast) ],$
or to the full state (that contains richer yet potentially distracting information): $Q^i(s_t, a_t) \leftarrow (1-\alpha ) Q^i(s_t,a_t)+ \alpha [r^i_t + \gamma Q^i(s_{t+1},a_t^\ast)].$
Here, $\alpha$ and $\gamma$ are hyper-parameters for Q-learning, while $a_t^\ast$ denotes the global optimal action. Further details, findings, and discussions are presented in the experiment section.

\subsubsection{Metric for sparsity and orthogonality}

\textbf{Sparsity.}
We consider the information shunt to be \textit{sufficient} in terms of reward recognition while being concise, such that any irrelevancy or redundancy information is masked out, resulting in a \textit{sparse} information flow. This property can be described as the maximization of \textit{information loss} after a state transformation $s_t \rightarrow \bar{\alpha}^i_t$.
That is, deducing the full state from the partial knowledge from a sub-state becomes more difficult as the information loss increases.
Following the definition from \citeauthor{DBLP:journals/corr/abs-1109-4856} (\citeyear{DBLP:journals/corr/abs-1109-4856}), the objective of maximizing the information loss for the $i$-th flow (i.e., transformation) is defined as 

\begin{align}\label{eq:l2}
    \begin{split}
        \max \sum_i \mathcal{L}(s_t \rightarrow \bar{\alpha}^i_t) \triangleq \max \sum_i \lim_{\hat{s}_t \rightarrow s_t} [ \mathcal{I}(\hat{s}_t; s_t) - \mathcal{I}(\hat{s}_t; \bar{\alpha}^i_t) ] = \max \sum_i \mathcal{H}(s_t|\bar{\alpha}^i_t),
    \end{split}
\end{align}

where $\mathcal{H}(s_t|\bar{\alpha}^i_t)$ is the conditional entropy indicating the uncertainty to deduce $s_t$ given $\bar{\alpha}^i_t$.

\noindent\textbf{Orthogonality.} To achieve that $\bar{\alpha}^i$ (or $\bar{\alpha}^j$) contains less or no information about determining $r^j$ (or $r^i$) (cf. Sec.~\ref{sec:noti-desid-expl}), we approximately regard this as the information orthogonality describing the independence between inter-states $\bar{\alpha}^i$ and $\bar{\alpha}^j$, which can be achieved by minimizing their mutual information, i.e., 
\begin{equation}\label{eq:l3}
    \min \sum_{i \neq j} \mathcal{I}(\bar{\alpha}^i_t; \bar{\alpha}^j_t).
\end{equation}




Note that the component reward $r^i$ can be given in advance (i.e., a known reward decomposition \citep{Juozapaitis19ExplainableReinforcement}) or be derived dynamically according to the distillation criteria \citep{lin2020rd}.
In the latter case for learning $\mathcal{R}$, explicit incentives for the consistency of $s^i$ and $r^i$ should be applied to avoid trivial solutions such as projecting all $K-1$ states to $0$ but leaving only one to $r$. 
For example, an objective of $\mathcal{I}(\bar{\alpha}^i; r^i)$ to maximize or $\mathcal{I}(\bar{\alpha}^i;r^j)$ to minimize when taking into account the orthogonality and the fact the $\bar{\alpha}^j$ should be aligned with $r^j$, but not $r^i$.

\subsection{Optimization procedure}


The overall optimization objective is a balanced combination of Eq.~\ref{eq:l0}, Eq.~\ref{eq:l1}, Eq.~\ref{eq:l2} and Eq.~\ref{eq:l3},
which involves neural estimation of entropy and mutual information \citep{Belghazi18MutualInformation, vandenOord18RepresentationLearning, lin2020rd, cheng2020club, Radford21LearningTransferable}.
For the estimation of mutual information, we individually approximate the entropy components\footnote{
Recall that $\mathcal{I}(X;Y) = \mathcal{H}(X) - \mathcal{H}(X|Y)$.
}
and follow previous work by \citeauthor{lin2020rd} (\citeyear{lin2020rd}) for the entropy approximation.
Future work will involve exploring the success of InfoNCE loss in contrastive learning \citep{vandenOord18RepresentationLearning} for better estimation.

In practice, considering the fact that $\bar{\alpha}^i$ is a subset of $s$, the knowledge of $s$ leads to the knowledge of $\bar{\alpha}^i$, such that
$\max \sum_i \mathcal{H}(s_t|\bar{\alpha}^i_t)
= \max \sum_i [\mathcal{H}(\bar{\alpha}^i_t|s_t) + \mathcal{H}(s_t)] - \mathcal{H}(\bar{\alpha}^i_t)
\approx \max \sum_i - \mathcal{H}(\bar{\alpha}^i_t)$,
which leads to an efficient estimation by, approximately, minimizing $\mathcal{H}(\bar{\alpha}^i)$.
This approximation reduces to the objective applied in previous works which can be optimized by minimizing one of its upper bounds in proportion to $\sum_i \log |m^i(s)|$ \citep{DBLP:journals/corr/abs-1109-4856, lin2020rd}.
We additionally optimize it with an $L_1$ penalty $\sum_i |m^i(s)|$ 
for the sake of sparse weights and stable information transition\footnote{The objective without ``$\log$'' can be derived from the perspective of $f$-mutual information but the choice of proper information measure, e.g., Kullback–Leibler or Jensen–Shannon divergence, remains undetermined for sophisticated learning systems.} 
\citep{li2023irrl}, and experimentally demonstrate its effectiveness.

We refer to the technique that employs the objectives (in Eq.~\ref{eq:l0}, Eq.~\ref{eq:l1}, Eq.~\ref{eq:l2}, Eq.~\ref{eq:l3}) to distil $\bar{\alpha}^i$ of a state, which in turn dictates the reward component $r^i$, as \textit{R-Mask}. In the masked Q-learning, the reward components are known a priori, and the acquisition of $\bar{\alpha}^i$ is synergized with the RL objective, alongside the objectives (in Eq.~\ref{eq:l0}, Eq.\ref{eq:l2} and Eq.\ref{eq:l3}) governing mask updates. This technique for mask learning is referred to as \textit{Q-Mask}.
As an ablation, their counterparts without sparsity and orthogonality losses (Eq.~\ref{eq:l2} and Eq.~\ref{eq:l3}) are denoted as \textit{R-Mask Lite} and \textit{Q-Mask Lite}, respectively.

\section{Experiments}\label{sec:experiment}

The following research questions outline the progressive evaluation of our explanation framework through extensive experiments:
\begin{enumerate}[label=Q\arabic*.]
  \setlength\itemsep{-0.4em}
  \item In comparison to vanilla RD \citep{Juozapaitis19ExplainableReinforcement}, how does the auxiliary task of decomposing reward (i.e., predicting $r^i(s, a)$) influence the generation of explanation artefacts? 
  \item Following reward prediction in Q1, what insights can be gained about the role of causal sufficiency of reward components (i.e., estimating $r^i(\bar{\alpha}^i, a)$ in Sec.~\ref{sec:me-ca-suff}) in learning causal factors? 
  \item Compared to the causal sufficiency of reward components above, how does the causal sufficiency concerning action (Sec.~\ref{sec:me-ca-suff}) impact the learning of causal factors \textit{uniquely}? 
  
\end{enumerate}

For a comprehensive list of methods and their distinctions studied in the experiments addressing the research questions, please refer to Table~\ref{table:all methods}.
An illustration of training flows of the R-Mask and the Q-Mask can be found in Fig.~\ref{fig:r-mask training flow} and Fig.~\ref{fig:q-mask training flow} in the Appendix, respectively. Neural network architecture details can be found in Appendix~\ref{det-neu-net-arc}, and pseudocode in Appendix~\ref{appendix:D}.

\subsection{Experimental setup}\label{sec:exp-setup}
To validate our causal attention principles in agent learning and answer the research questions, we conduct experiments on tasks of varying complexity and scale. 
We use two Atari 2600~\citep{DBLP:journals/corr/abs-1207-4708} tasks from OpenAI Gym~\citep{brockman2016openai}, including Gopher and MsPacman.

\textbf{Environments.} In the \textbf{Gopher} game ($K=2$), a farmer (i.e., the agent) protects carrots from a gopher. The agent receives a reward of 0.8 for bonking the gopher as it emerges from the holes or anywhere above ground, and a reward of 0.15 for filling those holes before the gopher tunnels out and eats carrots. In the \textbf{MsPacman} game ($K=3$), Pacman walks through a maze populated with various items (e.g., enemies and dots) and its object is to score as many as possible by eating them.
The multiple-reward structure in the game is as follows: the agent receives a reward of 0.25 when it gobbles a Dot up and a reward of 1 for eating an Energy Pill. When the agent gulps down one Energy Pill, the ghosts turn blue and Pacman can eat them. It earns a reward of 5 for each ghost (maximum 4 ghosts, i.e., 20) gobbled up. Note that we also introduce a \textbf{MiniGrid} toy task when addressing research question Q3.

\textbf{Performance.}
This metric represents the maximum score attained by the RL agent in a task. 

\textbf{Critical State.}
Given the human interest in understanding agent decisions relative to expectations, not all encountered states hold equal explanatory value. Critical states, characterized by significant utility gaps between optimal and second-best actions, are of particular interest. We evaluate and explain states as \textit{critical} based on the utility gap: $C(s) = \max_{a} Q(s, a) - \text{second-highest}_{a} Q(s, a)$, as specified in \cite{Amir2018HIGHLIGHTSSA, septon2023integrating}. 
We also consider states where the agent receives positive rewards (only non-negative rewards in the Atari tasks we considered).

\subsection{Analysis of research questions}\label{sec:ana-res-ques}


\subsubsection*{Q1. How does the task of reward prediction influence the explanation generation?}
\begin{figure}[ht] 
\centering
\begin{minipage}{0.3\textwidth}
\centering
\captionof{table}{Evaluation results for RD, RD-pred, RD-pred-u.}
\scalebox{0.72}{
\begin{tabular}{c|l|c}
\toprule
\multicolumn{2}{c|}{Evaluation Metric}&Performance \\
\midrule
\midrule

\multirow{3}{*}{Gopher} & RD & $\mathbf{15.62 \pm 1.58}$ \\
  & RD-pred & $14.43 \pm 0.41$ \\
& RD-pred-u &$13.78 \pm 0.21$  \\

\bottomrule
\end{tabular}
}
\label{table1: rd vs. rd-pred vs. rd-pred-u}
\end{minipage}%
\hfill
\begin{minipage}{0.65\textwidth}
    \centering
    \includegraphics[scale=0.46]{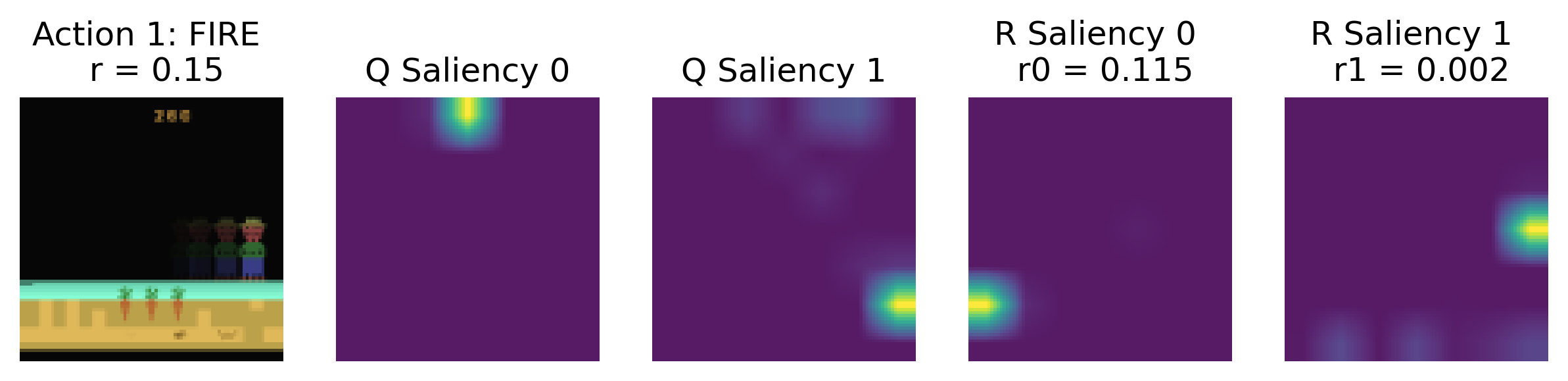}
    \captionof{figure}{Comparison of saliency maps (associated with ground and gopher rewards) of RD with RD-pred in a state where the agent filled the hole and attained reward 0.15. Q saliency refers to the generated saliency of Q-value; R saliency pertains to the generated saliency of reward.
      }
    \label{fig:Q1 rd vs. rd with reward prediction}
\end{minipage}%

\end{figure}

This question is raised under the hypothesis that our understanding of an agent's behaviour may benefit from probing other aspects (e.g., reward) of the agent's interaction data. Hence, on top of RD\footnote{Refer to Appendix~\ref{sec:dqn-rd} for a concise description of how RD functions. 
}, we introduce an auxiliary task where the agent learns to predict reward components $r^i(s, a)$, each supervised by a ground truth sub-reward signal. We denote this variant as \textit{RD-pred}.
Compared with RD in Table~\ref{table1: rd vs. rd-pred vs. rd-pred-u}, the performance drop is only considered moderate ($-7.62\%$).
However, we illustrate that the reward prediction task helps interpretability of agent behaviour.

To visually differentiate the resulting explanations\footnote{In vanilla RD, explanations typically involve sub-Q-value trade-offs or their differences.}, we adopt the GradCAM technique to generate post-hoc saliency maps for each component Q-value and reward $r^i$ concerning a state $s$.
Fig.~\ref{fig:Q1 rd vs. rd with reward prediction} shows that Q-value saliency associated with the ground reward erroneously focuses on the scoreboard, leading to a causal fallacy of putting the effect before the cause.
In contrast, R saliency attends to temporarily relevant, yet not precise areas  (e.g., leftmost ground and avatar body).
This can be attributed to the fact that predicted rewards reflect the value of transitioning to the next step from the current state, while Q-values reveal the expected long-term gain that may result in distortion of the causal structure because of this information compression along the time-axis.
This finding indicates that reward saliency is more informative in terms of interpreting the agent's temporary behaviour than Q-value saliency.

In the following section, we will introduce further learning objectives to explore causal structures (cf. Sec.~\ref{sec:noti-desid-expl}).



\subsubsection*{Q2. What is the gained insight into the role of causal sufficiency of reward components in learning causal factors in the R-Mask approach?}

\begin{figure}[!htb]
    \centering
    \subfigure[R-Mask masks for a state with reward $r = 0$.]{%
        \includegraphics[width=0.48\textwidth]{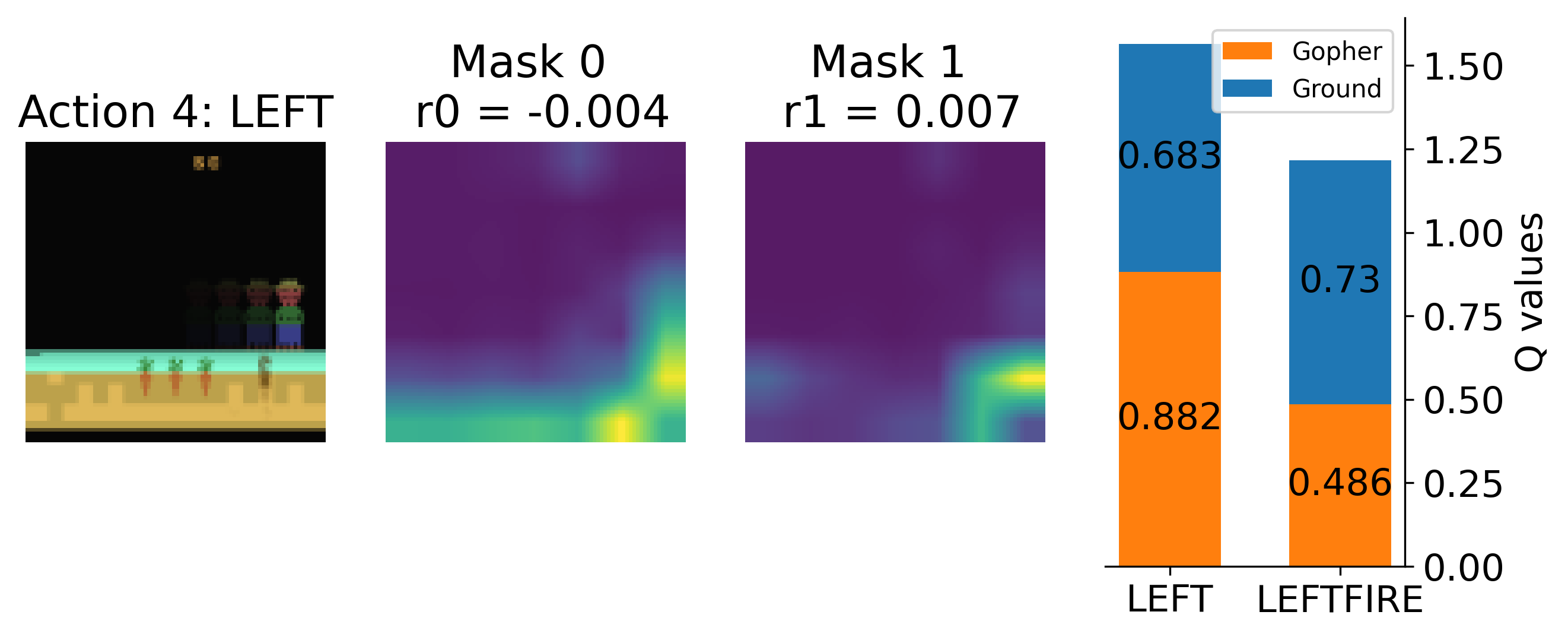}
        \label{fig:gopher masks with r=0.}
    }
    \hfill
    \subfigure[R-Mask masks for the \textbf{next} state with reward $r = 0.95$.]{%
        \includegraphics[width=0.48\textwidth]{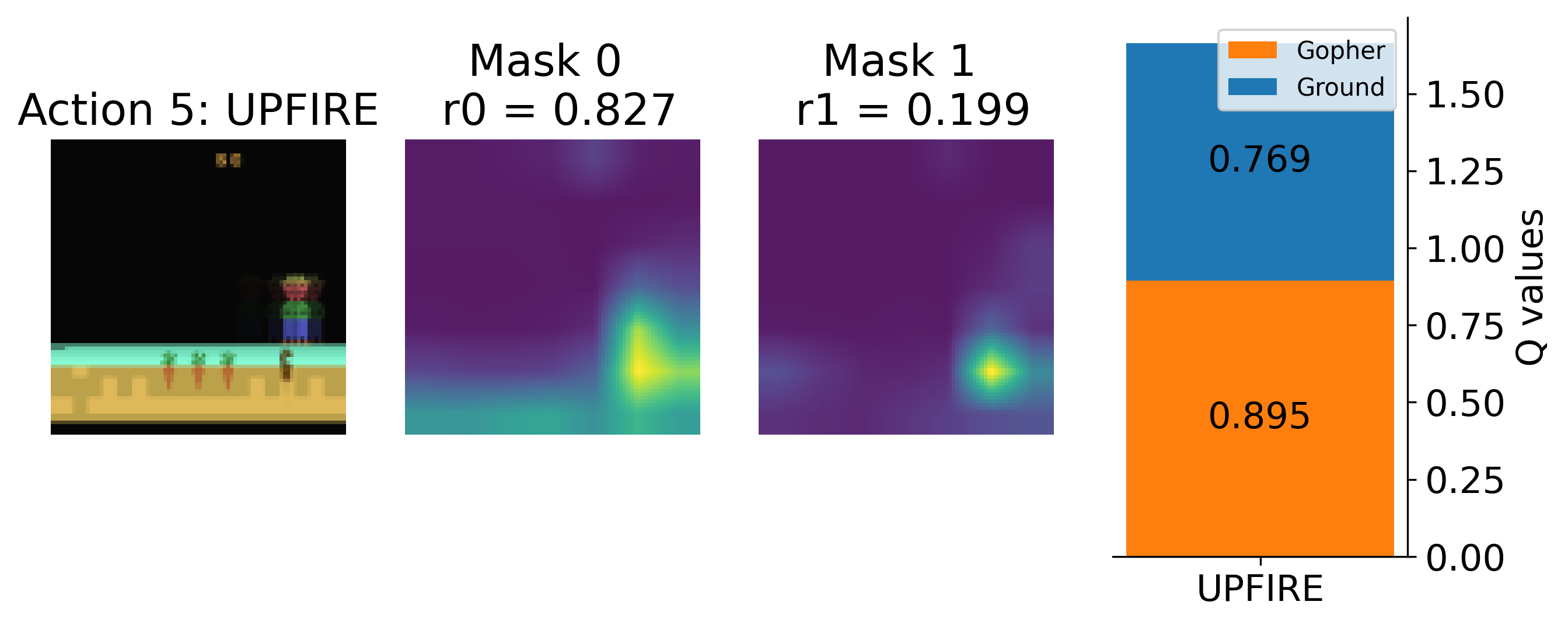}
        \label{fig:gopher masks2 with r=0.95}
    }
    \caption{R-Mask 
    attention masks\protect\footnotemark from Gopher and their interpretation along with Q-value bars. 
    }
    \label{fig:important gopher masks r-mask}
\end{figure}

The RD-pred approach, a variant of RD with reward prediction, does not encourage the information transition to be sufficient as a full state (i.e., all environmental aspects) is used to deduce $r^i$, thus complicating the disentanglement between reward components. The R-Mask approach constrains this information flow by employing the aforementioned objectives (Sec.~\ref{sec:learning framework}) to distil disentangled components of a state. Its effectiveness can be seen in Fig.~\ref{fig:important gopher masks r-mask} where causal factors (represented as attention masks) precisely identify relevant areas, enhancing our understanding of the agent's attention. For a fair comparison, we introduce a modified RD-pred with \textit{unknown} sub-rewards, denoted by \textit{RD-pred-u}, which only uses full reward supervision (for reward prediction) similar to R-Mask (see Table~\ref{table:all methods}). Masks generated by R-Mask emphasize more relevant objects, such as the avatar and gopher in Fig.~\ref{fig:important gopher masks r-mask}, while RD-pred-u focuses on irrelevant objects, like a flying bird, or loses focus entirely (as observed in Fig.~\ref{fig:gopher rd-pred-u mask1 with r=0.15} and Fig.~\ref{fig:gopher rd-pred-u mask2 with r=0.15} in the Appendix). This underscores the necessity of explicit signals (like those in Sec.~\ref{sec:learning framework} relied upon by R-Mask) to establish the correspondence between environmental aspects and sub-rewards. Interestingly, despite the performance drop in RD-pred-u (Table~\ref{table1: rd vs. rd-pred vs. rd-pred-u}), R-Mask achieves a relatively higher task return, albeit slightly lower than the baseline RD performance.

\footnotetext{Note, in Fig.~\ref{fig:Q1 rd vs. rd with reward prediction}, reward r0 signifies the ground reward as task specified. Here, r0 denotes the gopher reward, which we manually verify post-hoc after decomposition has been learned.}

\begin{figure}[!htb]
    \centering
    \subfigure[R-Mask masks for a state with reward $r=0$.]{%
        \includegraphics[width=0.49\textwidth]{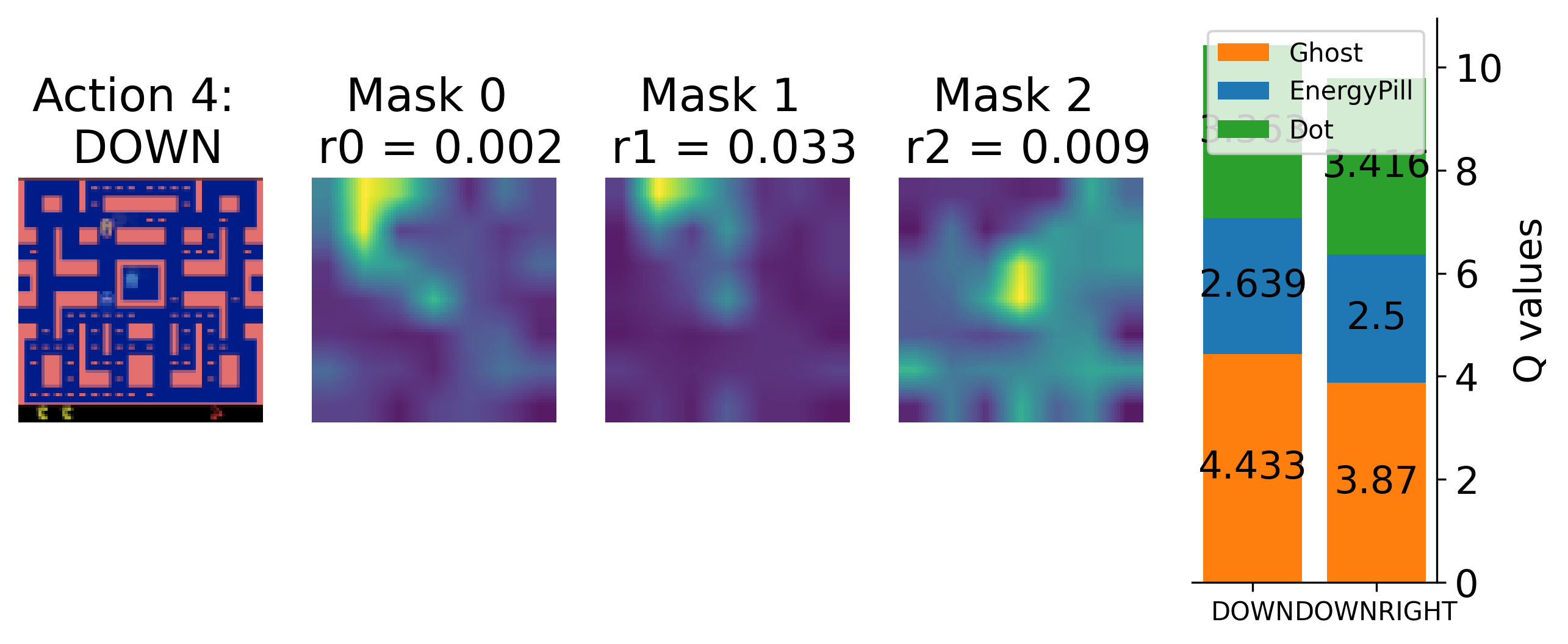}
        \label{fig:pacman masks r-mask main}
    }
    \hfill
    \subfigure[R-Mask masks for the \textbf{next} state with reward $r=5$.]{%
        \includegraphics[width=0.49\textwidth]{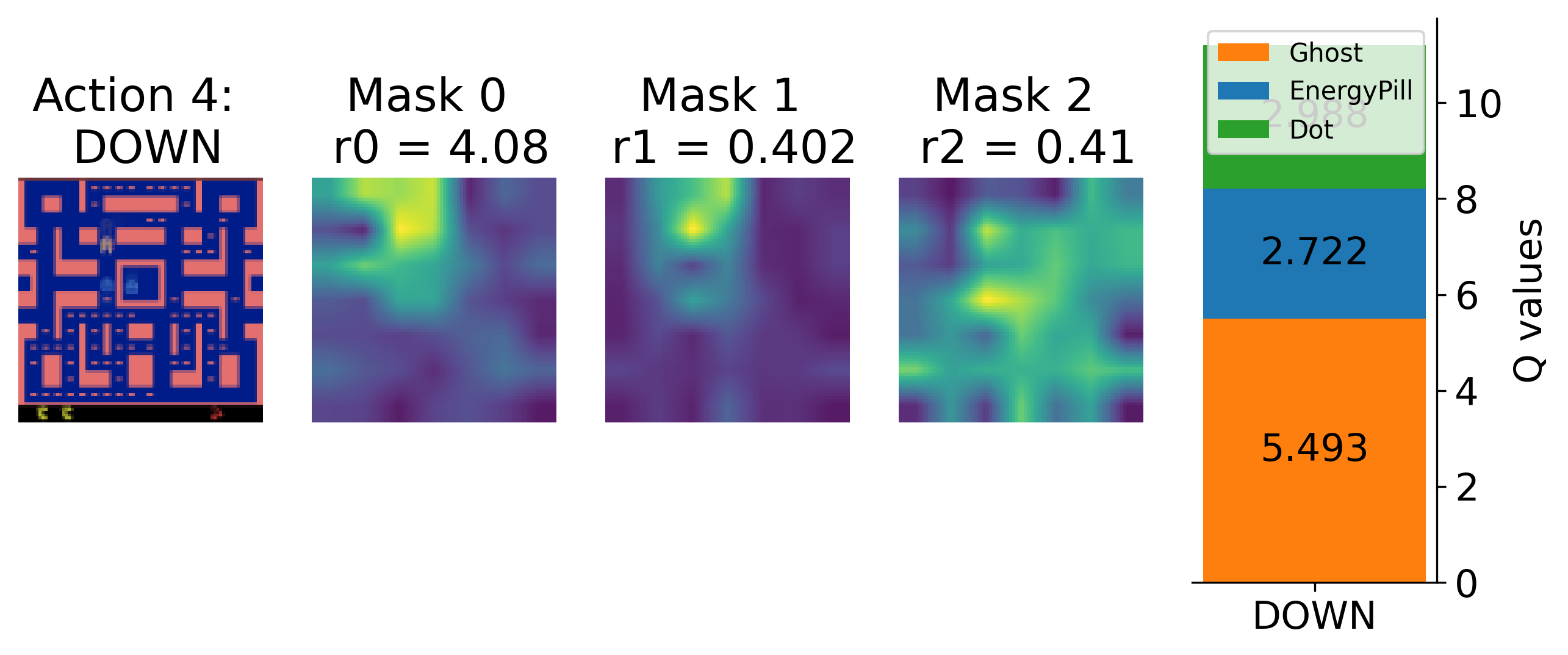}
        \label{fig:pacman masks r-mask at next state main}
    }
    \caption{R-Mask attention masks from MsPacman and their interpretation. (a) The masks (Mask 0 attends to ghosts, Mask 1 to energy pills, and Mask 2 to dots) and bar plots are for a critical but non-rewarding scenario. For a full description of the scene, please refer to Fig.~\ref{fig:important pacman masks r-mask} in Appendex~\ref{sec:add-re-pa-env}.
    }
    \label{fig:important pacman masks r-mask main}
\end{figure}

    
    
\begin{table}[th]
\centering
\caption{
Evaluations on Atari tasks. Metrics include fidelity (higher is better), sparsity (lower indicates sparser as desired), orthogonality (higher for better factor disentanglement), and task return performance.
}
\scalebox{0.6}{
\begin{tabular}{c|>{\centering\arraybackslash}p{2.0cm} p{2.0cm} p{2.3cm} p{2.3cm}|>{\centering\arraybackslash}p{2.0cm} p{2.0cm} p{2.3cm} p{2.3cm}}
\toprule
\multirow{2}{*}{\thead{Evaluation \\ Metrics}} & \multicolumn{4}{c|}{Gopher} & \multicolumn{4}{c}{MsPacman} \\
\cmidrule{2-9}
& Q-Mask  & Q-Mask Lite & R-Mask & R-Mask Lite & Q-Mask & Q-Mask Lite & R-Mask & R-Mask Lite \\
\midrule
\midrule

Fidelity & \NA& \NA & $\mathbf{84.58 \pm 0.64} \%$  &  $79.92 \pm 0.95 \%$ & \NA &  \NA & $ 65.75 \pm 0.85\%$ & $ \mathbf{88.16 \pm 0.09}\%$ \\ 
Sparsity & 0.782 & 0.468 &  0.106 &  0.488  &  3.4e-4  & 0.826 &$0.435$ & 0.932 \\
Orthogonality & -0.24 & 5.63 &  9.43  & 2.8 & 27.42  & 41.06 & $-8.449$ & 32.74 \\
Performance & $13.56 \pm 2.58$  & $12.17 \pm 3.06$ & $\mathbf{14.54 \pm 2.04}$  &  $12.48 \pm 0.83$  & $19.75 \pm 0.11$  & $\mathbf{29.94 \pm 0.16}$  & $27.86 \pm 0.59$ & $29.65 \pm 0.16$\\

\bottomrule
\end{tabular}}

\label{table: evaluation of attention masks for gopher and pacman}
\end{table}

\subsubsection*{Q3. How does the causal sufficiency concerning action impact the learning of causal factors uniquely?}\label{sec:research-question3}

The information sufficiency of determining the rewards and optimal decisions for an agent are highly correlated but not necessarily equivalent.
The agent from the Q-Mask consumes the distilled state (i.e., factors) and insufficient factors may exert a challenge in optimizing a policy, which may stem from many factors such as unstable Q-agent update. Thus, finding an appropriate disentanglement is deemed not straightforward in this case. The lower task return in Table~\ref{table: evaluation of attention masks for gopher and pacman} evidence our first observation.

We further propose more tractable and human-intuitive evaluation metrics to quantitatively gauge the attainment of the desired behaviour of masks.
\textit{Fidelity} computes as $\frac{\#(a^*=\hat{a}^*)}{\#(a^*)}$, measuring the consistency of decision $a^{*}$ made with full state and the decision $\hat{a}^{*}$ with distilled state.
\textit{Sparsity} roughly measures the decrease of the information capacity (the lower the better) when the state is masked, computed as $\frac{|\bar{\alpha}^i|}{|s|}$.
Finally, to approximately measure state inter-independency, we count the overlap of masks regarding \textit{orthogonality}.
(See Appendix~\ref{sec:eval-metrics} for derivation and explanation.)

Comparing R-Mask masks in Fig.~\ref{fig:important gopher masks r-mask} and Q-Mask masks in Fig.~\ref{fig:gopher mask q comparison} (rightmost two columns), though both deliver us a visual intuition that R-Mask attention masks are more distinct, more orthogonal, and void of spurious objects~\citep{DBLP:journals/corr/abs-1906-11883, Wu2021SelfSupervisedAR}. One explanation is that top-down attention (e.g., Q-Mask) is guided explicitly by the RL objective. As a result, the mask shaping becomes heavily tied to this objective, potentially causing the agent to link its rewarding behaviour with changes in displayed scores.
This, in turn, can inevitably introduce bias in the causal relationship between state representation and chosen actions.

On the hypothesis that challenging tasks, especially the ones with a high-dimensional state, usually lead to unstable training and thus difficulties of distillation of causal factors, we further conduct experiments on a toy task, Monster-Treasure~\citep{gym_minigrid}, where the ground truth of causal factors are accessible and manageable for analysis. It turns out that, on tasks with low-dimensional states and easily disentangable causals, the Q-Mask shows better alignment with the ground truth than the R-Mask, which indicates that feeding the agent with distilled states helps both, reward prediction and state disentanglement. See Appendix~\ref{sec:monstercase} for detailed case analysis.






Notably, while no definitive benchmarks exist for optimal orthogonality and sparseness, lower values are preferable, i.e., disentangled and sparse factors are favoured. In the process of learning masks, there exists a trade-off between sparsity and orthogonality. When contrasting evaluation results in Table \ref{table: evaluation of attention masks for gopher and pacman} within the Gopher context, a notable trend emerges: a lower level of sparsity tends to correlate with a heightened degree of orthogonality. However, for the Pacman task, we observe the opposite pattern. Not surprisingly, depending on the specific RL task, the optimal balance of those desiderata may vary. Nevertheless, these indicators generally align with our perception of the generated explanations and prevent trivial and irrelevant causal factors from being learned. A comprehensive description of how the proposed desiderata contribute to our understanding of the agent's behaviour can be found in Appendix~\ref{sec:com-rm-qm-lite}.
See case studies for details in Sec.~\ref{sec:case studies} below.

\subsection{Case studies}\label{sec:case studies}



\textbf{R-Mask Attention Masks on Gopher.} 
We showcase attention masks learned by R-Mask in a critical scenario (Fig.~\ref{fig:important gopher masks r-mask}). The agent's preference for the ``LEFT'' move over ``LEFTFIRE'' in a critical scenario is explained by a larger Q-value difference under the gopher reward component (see computation in Appendix~\ref{sec:details-com-reward-de-ex}). This indicates that the agent is aiming for double rewards by moving left before executing a ``UPFIRE'' action when the gopher emerges, as supported by the analysis of attention masks provided by R-Mask (e.g., as the agent nears the object, Mask 0 and Mask 1 follow and contract). Note that attention masks adeptly capture subtle nuances in the two visually similar scenarios, which is crucial for understanding the agent's one-step action. Furthermore, the R-Mask method accurately predicts reward components in the scene, bolstering our confidence in explaining the agent's preference for ``LEFTFIRE'' through R-Mask's attention masks. For an in-depth case study, please refer to the Appendix~\ref{sec:case-studies-appendix}.

\noindent \textbf{R-Mask Attention Masks on MsPacman.}
To further validate the ability of the proposed methods to mine the cause-effect relationships for more challenging environments when the reward causes are actually \textit{interdependent}, we test R-Mask on the MsPacman environment (Q-Mask results are in Appendix~\ref{sec:add-re-pa-env}). The results in Fig.~\ref{fig:important pacman masks r-mask main} indicate that the method can reveal the agent's decision-making rationale in challenging scenarios, but there are challenges when rewards are interdependent, affecting the accuracy of reward prediction. A more detailed explanation of this can be found in Appendix~\ref{sec:case-studies-appendix}.





\section{Discussion}


In this paper, we present a novel approach to unravelling the complex relationships between model predictions, the reasoning mechanism, and explanations in reinforcement learning. On top of the non-post-hoc RD approach, we introduce a causal model that identifies explanatory factors contributing to an agent's decisions, which differs from traditional saliency-based methods. The proposed framework provides a diverse perspective on the agent's interactions and can be integrated with policy-level explanations, such as that by~\citet{NEURIPS2021_65c89f5a}, to identify critical time steps and localize features for a deeper understanding of the agent's attention history.

\textbf{Limitations.} Our approach assumes the existence of multiple channels, which might not always hold. Challenges may arise when rewards are interdependent or tasks involve numerous reward components, potentially affecting computational efficiency. Achieving full invariance of factors through intervention to irrelevant task components may not always be feasible, particularly in complex tasks.

\textbf{Outlook.} Although we focus on the use of learned causal factors to generate explanations by visualizing factors, represented by various masks, the learned factors can also be used to generate counterfactual explanations --- minimal perturbations of causal factors that change the agent's behaviour~\citep{DBLP:journals/corr/abs-2101-12446}. 
Another promising but challenging future direction is relaxing the assumption of multiple rewards, i.e., exploring a more general setting without sub-rewards. This introduces a more complex expression for information flow, i.e., how various causal factors contribute to a \textit{single} reward, but the same guiding desiderata would apply with some adjustments. The main challenge is assigning nontrivial meanings to factors when there exists one reward facet.
However, learning causal factors could be enhanced through the auxiliary task of modelling dynamics, i.e., by utilizing environmental changes as extra supervision, learned factors may be more interpretable. Finally, techniques like LLMs, which can convey the aspects controlled by each factor to humans in language, would further improve explanation quality.
\acks{This research was funded by the Federal Ministry for Economic Affairs and Climate Action (BMWK) under the Federal Aviation Research Programme (LuFO), Projekt VeriKAS (20X1905)}


\bibliography{references}

\newpage
\appendix

\section{Additional Results in Monster-Treasure Environment} \label{sec:ad-re-mt-env}

\subsection{Reward Estimate} Table \ref{tab:reward prediction with R-Mask} is referenced in the Experiments section (Sec.~\ref{sec:ana-res-ques}). The table documents the reward estimate corresponding to the state depicted in Fig.~\ref{fig:mt_example_full}.

\begin{table}[ht]
    \centering
    \caption{Reward Predictions with the R-Mask }
    \begin{tabular}{l|c|c|c|c}
        \toprule
        & Right & Down & Left & Up \\
        \midrule
        \midrule
        $r^0$ & 2.288 & 0.28 & 0.287 & 0.312 \\
        $r^1$ & -0.29 & -0.262 & -2.189 & -0.295 \\
        sum & 1.998 & 0.018 & -1.902 & 0.017 \\
        \bottomrule
    \end{tabular}
    \label{tab:reward prediction with R-Mask}
\end{table}

\subsection{Mask Scores} 
Given our knowledge of the ground truth masks in this environment, we depart from the metrics detailed in the Evaluation section (Sec.~\ref{sec:eval-metrics}). Instead, we capture the environment-specific mask score in Table~\ref{tab:mt_mask_scores}. This score quantifies deviation from the ideal masks for this setting: one concealing monster information (i.e., coordinates) and another hiding treasure details. A lower score indicates better masks, with scores below 1 signifying effective masks. 

\begin{table}[ht]
    \centering
    \caption{Mask Scores for Monster-Treasure Environment} 
    \begin{tabular}{l|c|c}
        \toprule
        & Mean & Standard Deviation \\
        \midrule
        \midrule
        Q-Mask & 0.507 & 0.302  \\
        R-Mask & 1.913 & 1.133 \\
        \bottomrule
    \end{tabular}
    \label{tab:mt_mask_scores}
\end{table}

\subsection{Performance and Mask Accuracy Trade-off} 
The average return for Q-Mask stands at 1.97, contrasting with R-Mask's value of 2. Despite Q-Mask's precise mask generation, its performance has slightly declined compared to R-Mask. This observation can be attributed to the fact that all Q-agents within Q-Mask are exposed solely to a partial environmental view generated by learnable mask networks (e.g., updated by Eq.~\ref{eq:l2} and Eq.~\ref{eq:l3}). Consequently, during the initial stages of mask learning, Q-agents might grapple with acquiring task-solving skills. This struggle could inadvertently lead to the erroneous filtering of both irrelevant and relevant information, possibly affecting task performance.


\section{Additional Results in Atari Environments} \label{sec:add-re-at-env}

\subsection{Additional Results in Gopher}\label{sec:add-re-go-env}

\begin{figure}[!htb]
    \centering
    \subfigure[]{%
        \includegraphics[width=0.83\textwidth]{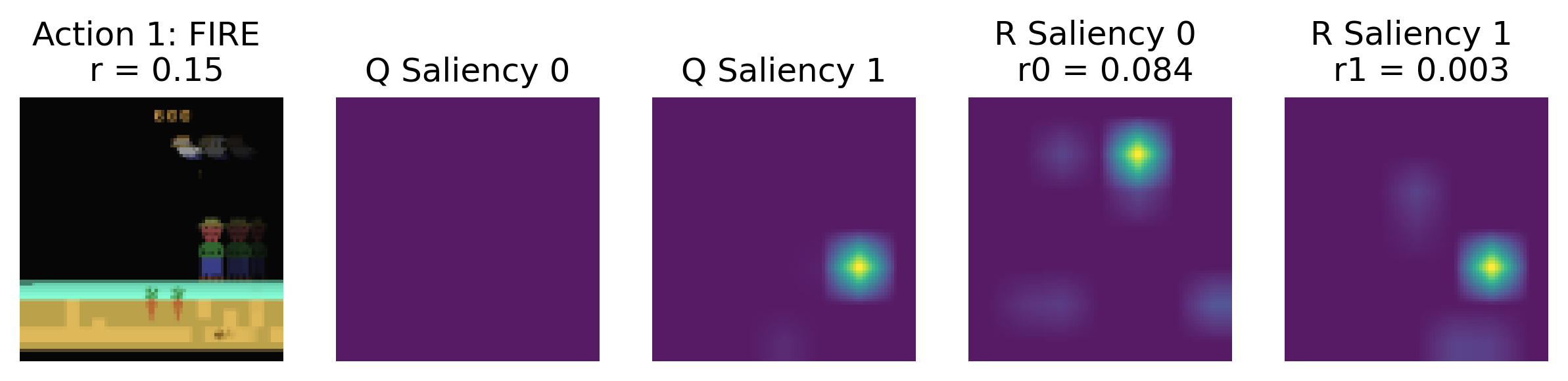}
        \label{fig:gopher rd-pred-u mask1 with r=0.15}
    }
    \hfill
    \subfigure[]{%
        \includegraphics[width=0.83\textwidth]{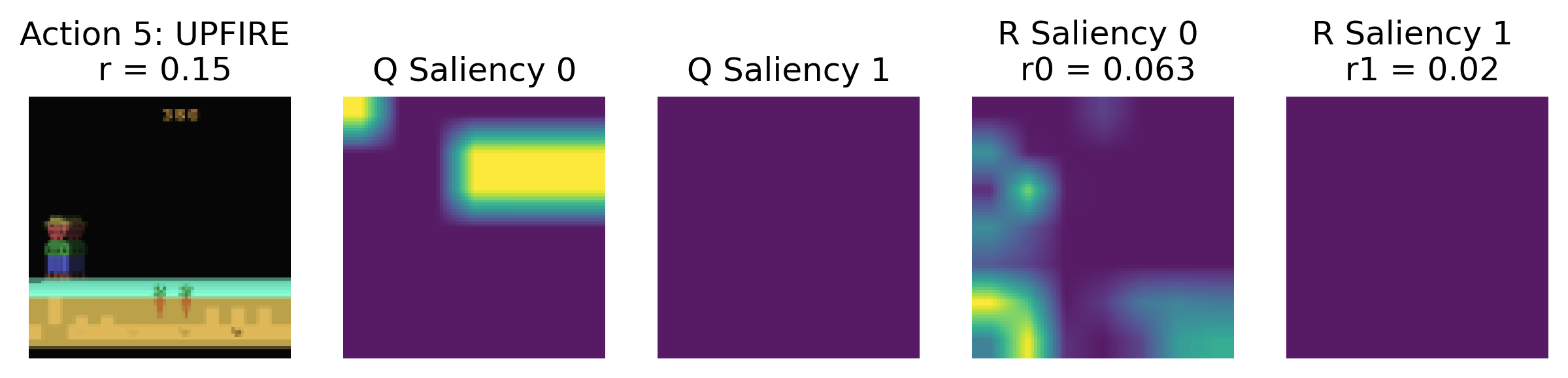}
        \label{fig:gopher rd-pred-u mask2 with r=0.15}
    }
    \caption{Comparison of saliency maps (associated with ground and gopher rewards) of RD with RD-pred-u in a state where the agent filled the hole and attained reward 0.15. 
      }
    \label{fig:Q2 rd-pred-u masks}
\end{figure}
\begin{figure}[!htb] 
      \centering
      \includegraphics[scale=0.78]{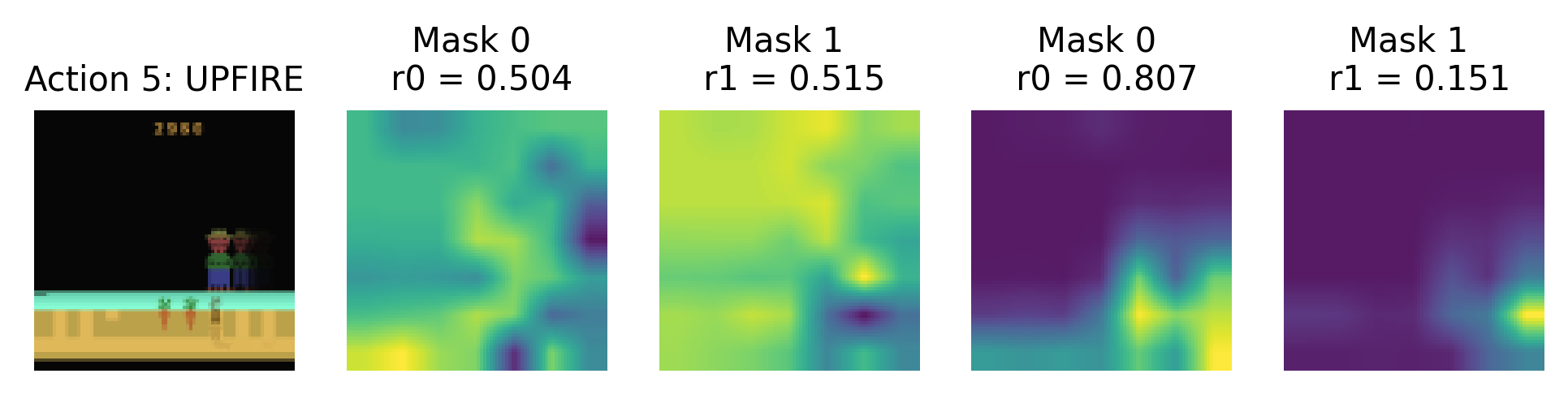}
      \caption{R-Mask Lite masks vs. R-Mask masks for a state with reward $r=0.95$.
      }
      \label{fig:gopher mask r-mask comparison 095}
\end{figure}
\begin{figure}[!htb] 
      \centering
      \includegraphics[scale=0.78] 
    {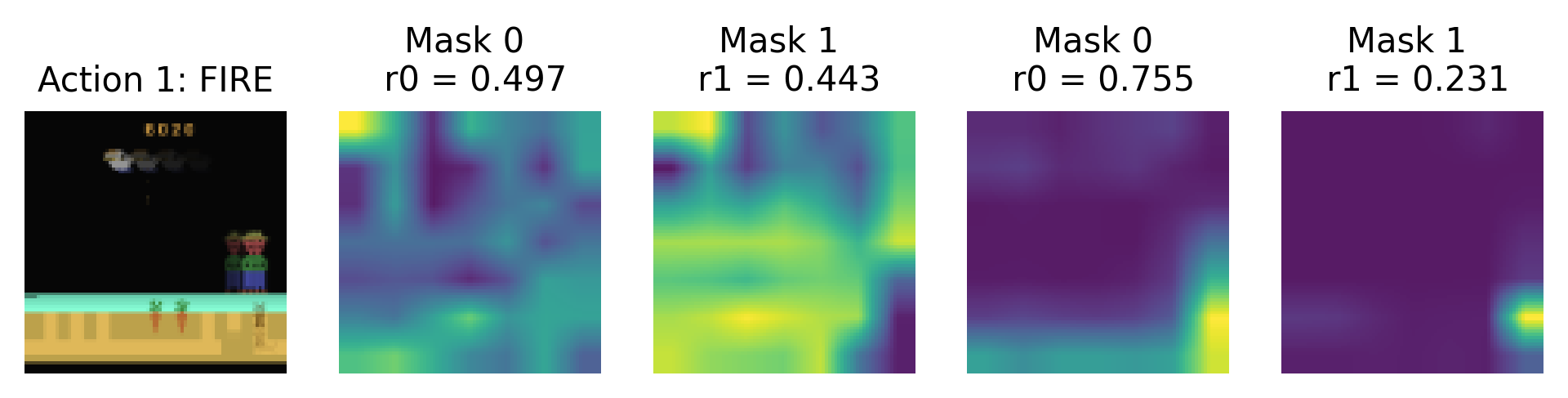}  
      \caption{R-Mask Lite masks vs. R-Mask masks for a state with reward $r=0.95$ (another example state where a flying bird recently passed by).
      }
      \label{fig:gopher mask r-mask comparison2 095}
\end{figure}
\begin{figure}[!htb]
      \centering
      \includegraphics[scale=0.78]{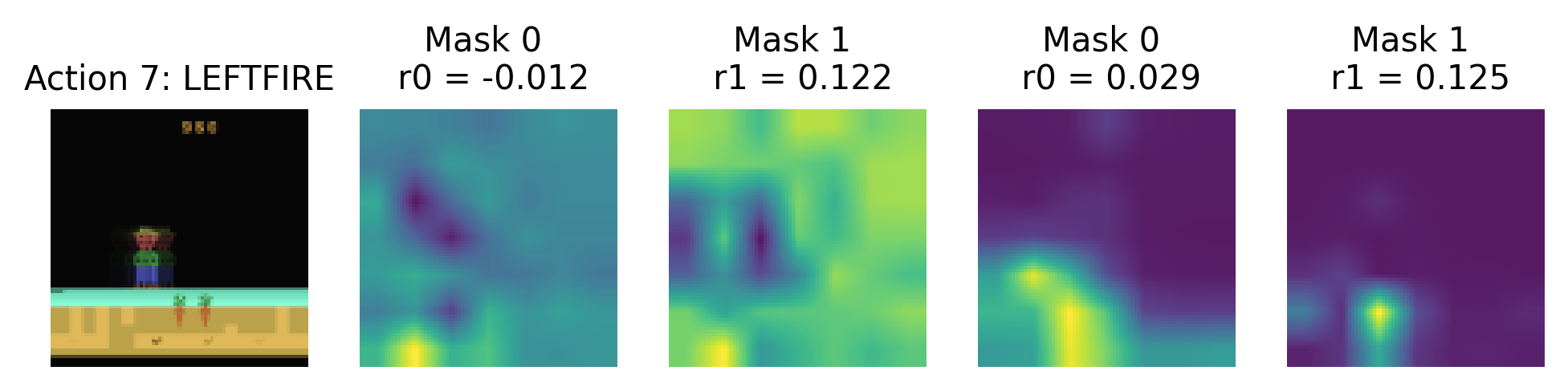}
      \caption{R-Mask Lite vs. R-Mask for a state with reward $r=0.15$.
      }
      \label{fig:gopher mask rd2 comparison 015}
\end{figure}
\begin{figure}[thpb]
      \centering
      \includegraphics[scale=0.67]{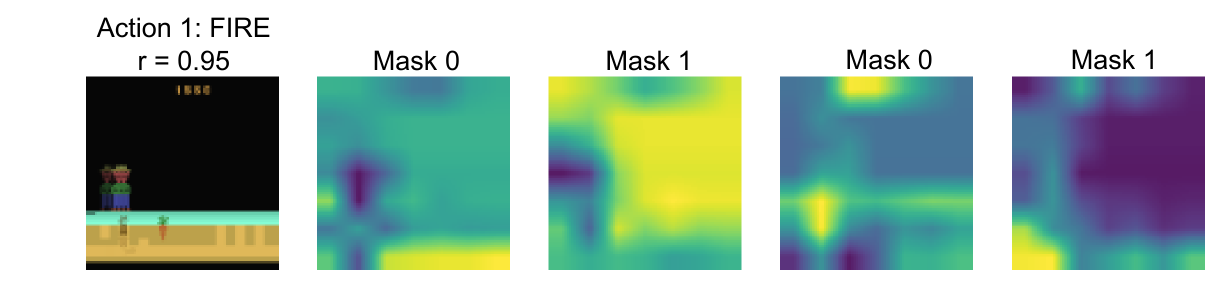}
      \caption{This figure depicts a rewarding state ($r = 0.95$), along with masks from Q-Mask Lite (first two) and Q-Mask (last two). 
      }
      \label{fig:gopher mask q comparison}
\end{figure}
\begin{figure}[!htb]
    \centering

    \subfigure[Q-Mask masks for a state]{%
        \includegraphics[width=0.88\textwidth]{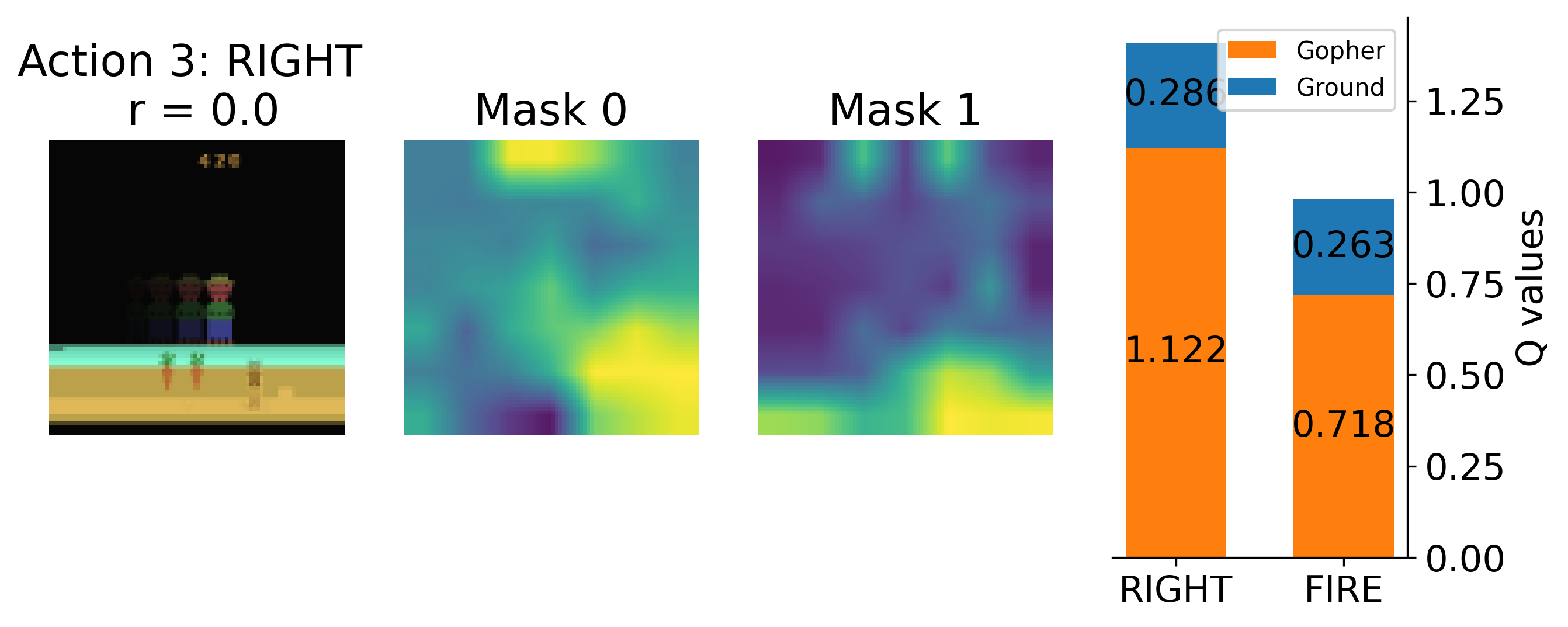}
        \label{fig:gopher masks q}
    }
    \hfill
    \subfigure[Q-Mask masks for the \textbf{next} state]{%
        \includegraphics[width=0.88\textwidth]{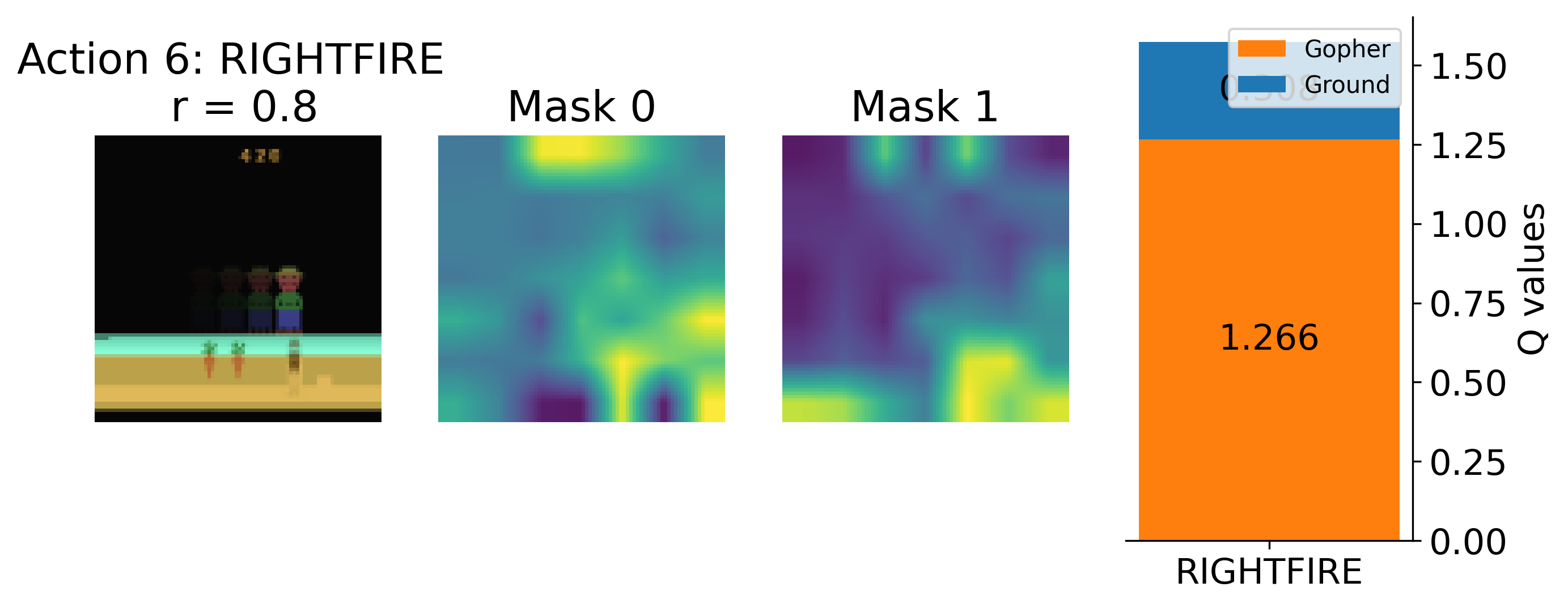}
        \label{fig:gopher masks q at next state}
    }
    
    
    
    \caption{Q-Mask attention masks from Gopher and their interpretation. (a) The masks (Mask 0 represents attention to the gopher while Mask 1 to ground) and bar plots are for a scenario (critical state with no reward), where there is a large Q-value gap between a chosen ``RIGHT'' move and a second-best ``FIRE'' action. The agent's choice to opt for a ``RIGHT'' move rather than a ``FIRE'' action as the gopher emerges from its hole is visually unclear. However, a closer examination of the following state (\ref{fig:gopher masks q at next state}) and the contracting attention masks (particularly areas at the bottom-right) exposes the gopher's strategy. It plans to ``RIGHTFIRE'' after moving right, intentionally aiming for a collision and a reward.}
    \label{fig:important gopher masks q-mask}
\end{figure}

Fig.~\ref{fig:Q2 rd-pred-u masks} is discussed in Sec.~\ref{sec:ana-res-ques}. We present examples that compare R-Mask (Q-Mask) with its lite variant in Figures~\ref{fig:gopher mask r-mask comparison 095}, \ref{fig:gopher mask r-mask comparison2 095}, \ref{fig:gopher mask rd2 comparison 015}, and \ref{fig:gopher mask q comparison}. Additionally, Fig.~\ref{fig:important gopher masks q-mask} presents an in-depth illustration of Q-Mask attention masks for the Gopher environment.

\FloatBarrier

\subsection{Additional Results in MsPacman} \label{sec:add-re-pa-env}
\begin{figure}[!htb]
    \centering
    \subfigure[R-Mask masks for a state with reward $r=0$.]{%
        \includegraphics[width=\textwidth]{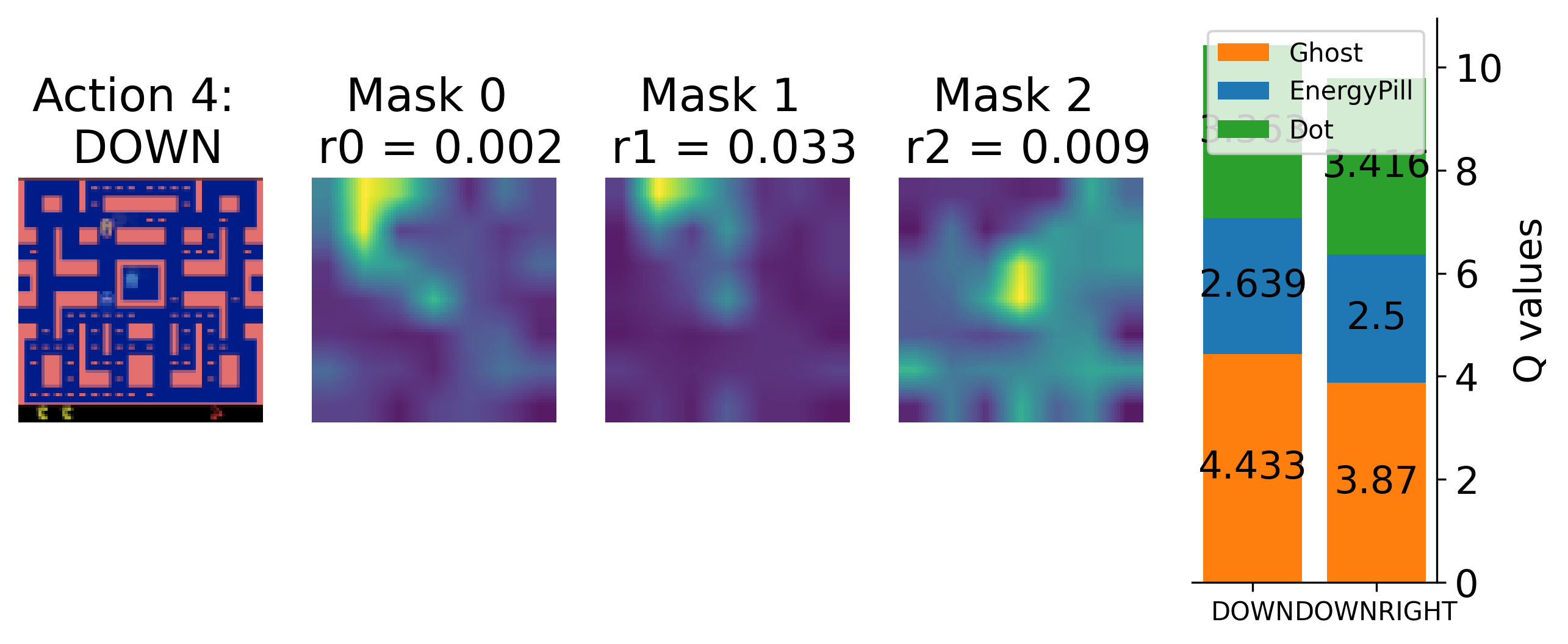}
        \label{fig:pacman masks r-mask}
    }
    \subfigure[R-Mask masks for the \textbf{next} state with reward $r=5$.]{%
        \includegraphics[width=\textwidth]{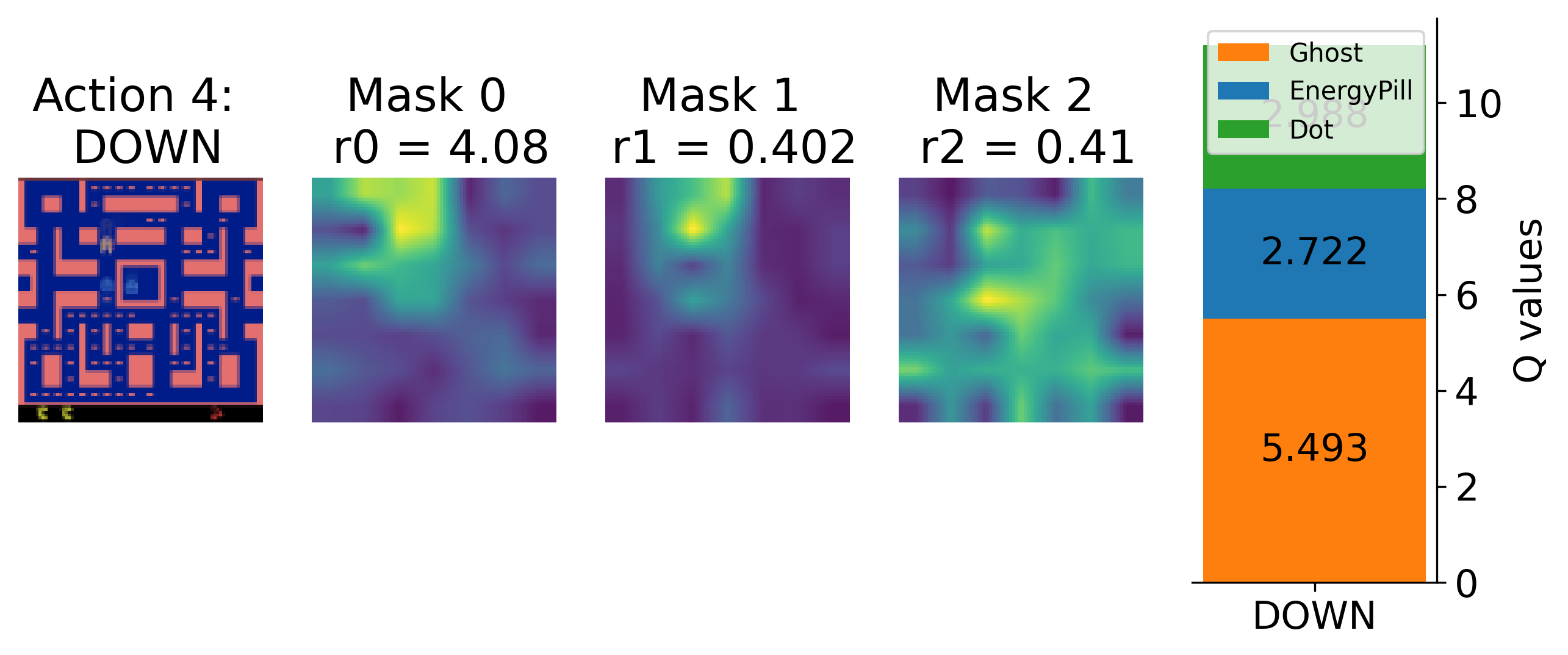}
        \label{fig:pacman masks r-mask at next state}
    }
    
    \caption{R-Mask attention masks from MsPacman and their interpretation. (a) The masks (Mask 0 attends to ghosts, Mask 1 to energy pills, and Mask 2 to dots) and bar plots are for a critical but non-rewarding scenario. Positioned at the top-left crossroad of the maze, the Pacman faces an imminent encounter with a ghost. In this state (Fig. \ref{fig:pacman masks r-mask}), the agent can select a ``DOWN'' move instead of a risky ``DOWNRIGHT'' action, evading the ghost. By examining the subsequent state and attention masks (\ref{fig:pacman masks r-mask at next state}), particularly the upper-left region, the Pacman's intention becomes evident. Detecting the ghost, the Pacman executes a ``DOWN'' move, causing a collision and thereby yielding a reward.}
    \label{fig:important pacman masks r-mask}
\end{figure}

\begin{figure}[!htb]
    \centering

    \subfigure[R-Mask masks for a state with reward $r=0$.]{%
        \includegraphics[width=\textwidth]{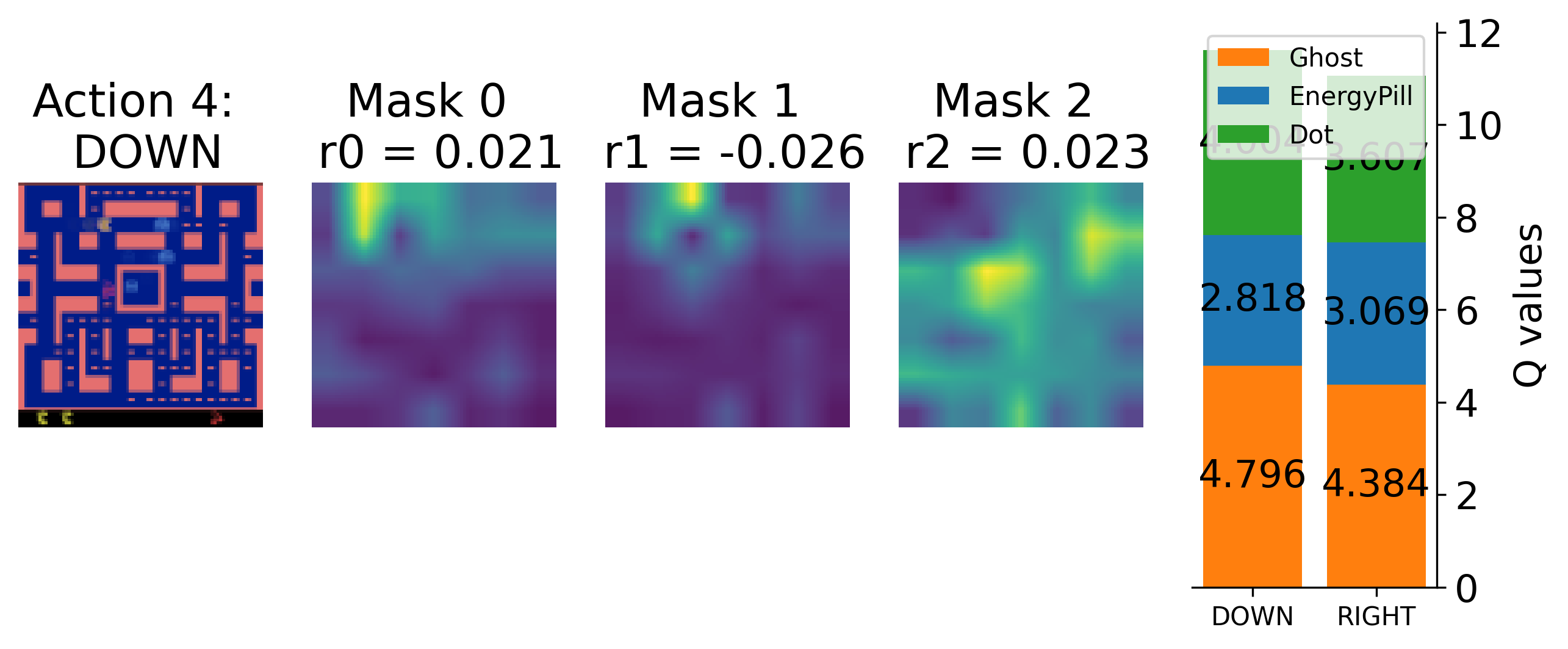}
        \label{fig:pacman masks2 r-mask}
    }
    \subfigure[R-Mask masks for the \textbf{next} state with reward $r=5$.]{%
        \includegraphics[width=\textwidth]{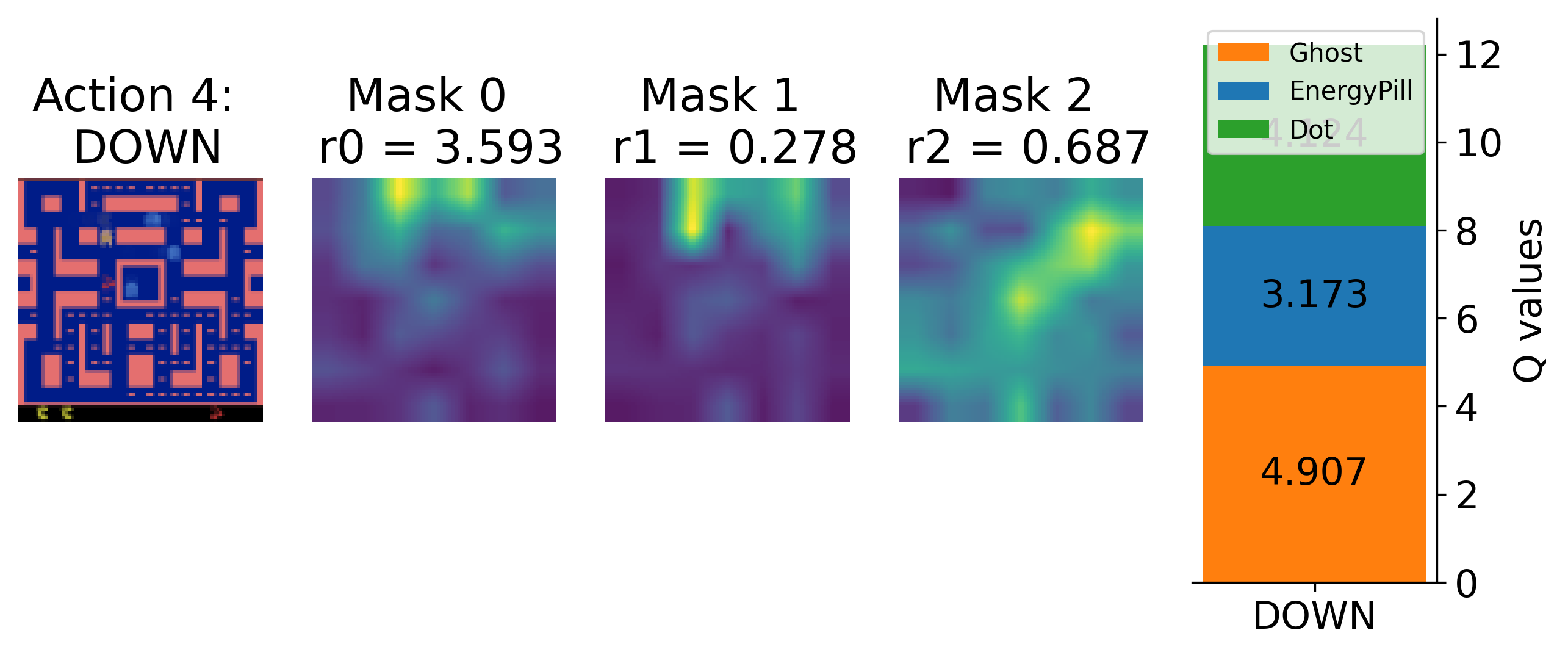}
        \label{fig:pacman masks2 r-mask at next state}
    }

    \caption{Another R-Mask mask in MsPacman environment.}
    \label{fig:important pacman masks2 r-mask}
\end{figure}

\begin{figure}[!htb]
    \centering
    \subfigure[Q-Mask masks for a state.]{%
        \includegraphics[width=\textwidth]{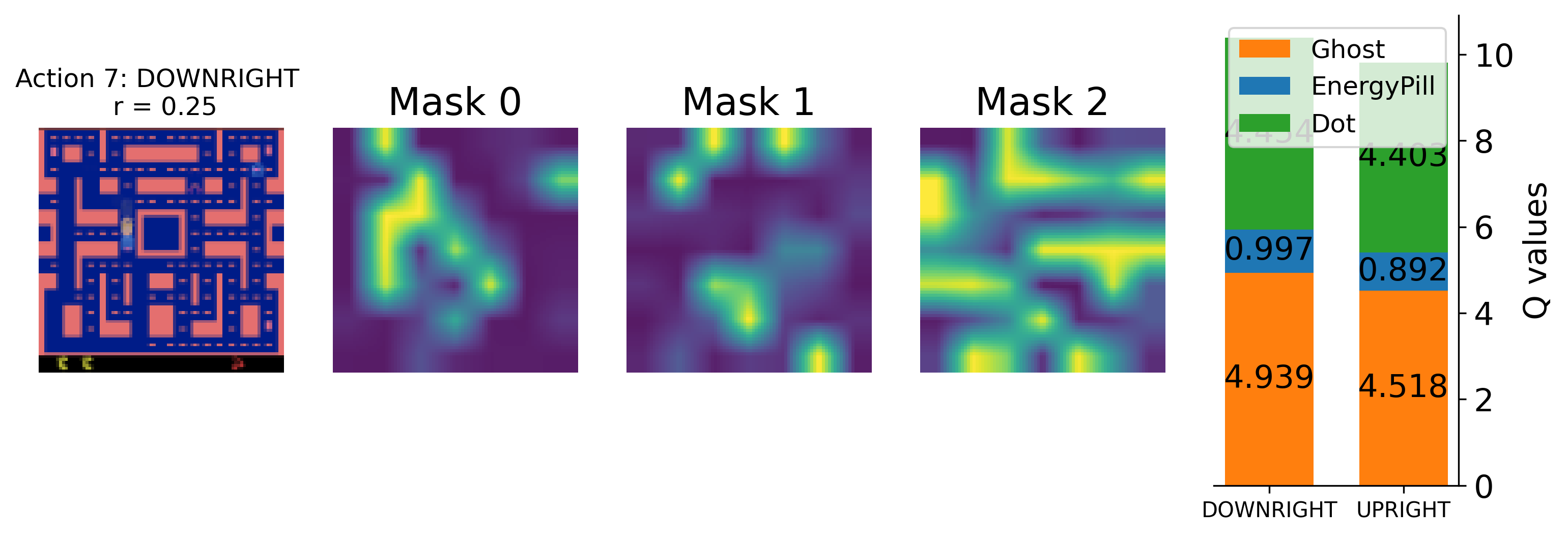}
        \label{fig:pacman masks q-mask}
    }
    \subfigure[Q-Mask masks for the \textbf{next} state.]{%
        \includegraphics[width=\textwidth]{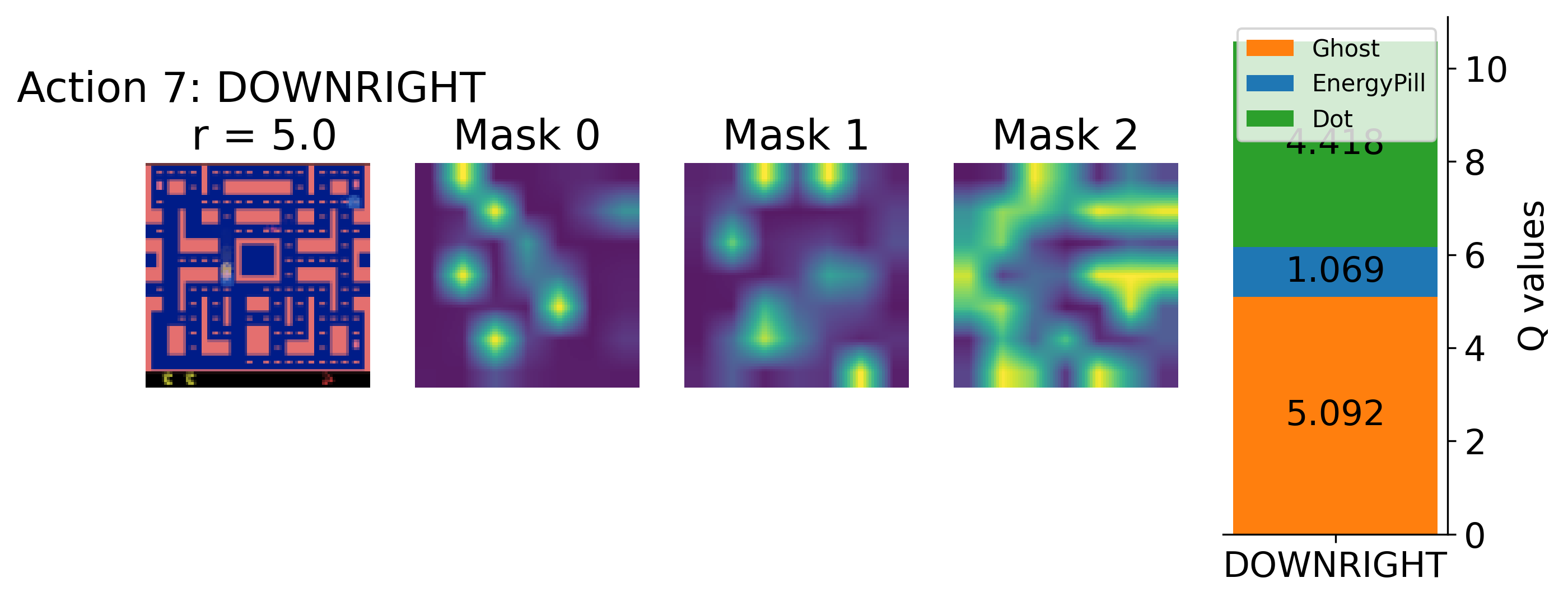}
        \label{fig:pacman masks q-mask at next state}
    }
    
    \caption{Q-Mask attention masks from MsPacman and
their interpretation. (a) The masks (Mask 0 attends to ghosts, Mask 1 to energy pills, and Mask 2 to dots) and bar plots are for a critical and rewarding scenario. As Pacman progresses downward within the middle-left maze area, it consumes a dot while simultaneously encountering a ghost. In this situation (depicted in Fig. \ref{fig:pacman masks q-mask}), the agent selects a ``DOWNRIGHT'' move over an ``UPRIGHT'' action, which would involve passing the ghost. An analysis of the subsequent state and attention masks (\ref{fig:pacman masks q-mask at next state}) exposes the Pacman's strategy. Recognizing the ghost, the Pacman continues its downward movement, resulting in a collision with the ghost and the subsequent reward.}
    \label{fig:important pacman masks q-mask}
\end{figure}

Fig.~\ref{fig:important pacman masks r-mask} is referenced in the Evaluation section (Sec.~\ref{sec:case studies}) when introducing R-Mask attention masks in the MsPacman environment. Furthermore, Fig.~\ref{fig:important pacman masks2 r-mask} presents another illustrative example of R-Mask masks designed for the MsPacman environment. In addition to these, we showcase an instance of Q-Mask masks in Fig.~\ref{fig:important pacman masks q-mask}.

\FloatBarrier

\subsection{Convergence in Agent's Learning}\label{con-in-age-lea}
As a reference for convergence (of episodic scores/returns for Atari environments), we put the statistics of scores (after 10M environment steps) in Table~\ref{tab:score-conver-diff-atar-ga}, in which we compare human baseline, common DQN approach, vanilla RD, and our approaches (extensions of RD) for games Gopher, MsPacman.
Note this table may only serve as a rough comparison as some methods are evaluated under different conditions, e.g., different neural network architectures, hyperparameters, and learning steps. Note, $\text{DDQN}^\ast$ and $\text{RD}^{\ast}$ represent our implementations of the DDQN and RD algorithms, respectively.

\begin{table}[th]
    \centering
    \caption{Scores at Convergence in Different Atari Games}
    \scalebox{0.8}{\begin{tabular}{l|c|c|c|c|c|c}
        \toprule
        \textbf{Games} & \thead{\textbf{Average Human} \\ \cite{DBLP:journals/corr/abs-2003-13350}}  & \thead{\textbf{DDQN} \\ \cite{DBLP:journals/corr/HasseltGS15}} & $\textbf{DDQN}^\ast$ & $\textbf{RD}^{\ast}$ & \textbf{R-Mask (our)} & \textbf{Q-Mask (our)} \\
        \midrule
        \midrule
        Gopher & 2412.5 & 8742.8 & 8338 & 7881.3 & 8671.3 & 8078.8 \\
        MsPacman & 6951.6 & 1401.8 & 6132 & 5810 & 6961.8 & 5818.4 \\
        \bottomrule
    \end{tabular}}
    \label{tab:score-conver-diff-atar-ga}
\end{table}

\section{Full Reference to Main Text} \label{appendix:C}

\subsection{Formalization of RL Problem with SCM}\label{sec:forma-rl-scm}
We formalize the RL problem with the following structural causal model (SCM):

\begin{equation}
    \mathcal{S} := f_{\mathcal{S}}(\alpha, \beta, U_{\mathcal{S}}),\, \mathcal{A} := f_{\mathcal{A}}(\alpha, U_{\mathcal{A}}),\, \mathcal{R} := f_{\mathcal{R}}(\alpha, \mathcal{A}, U_{\mathcal{R}}),
\end{equation}
where noise variables are jointly independent: $U_{\mathcal{S}} \perp U_{\mathcal{A}} \perp U_{\mathcal{R}}$. As for $f_{\mathcal{S}}, f_{\mathcal{A}}, f_{\mathcal{R}}$, they are unknown structural functions; $f_\mathcal{A}$ can be regarded as the policy to be learned and causal factors $\alpha$ can be obtained by a masker $m(\cdot)$ which we will detail in the main text.

\subsection{Computing Causal Intervention}\label{sec:computing-ca-suff}
Formally, given an environment state $s$, its Fourier transformation is expressed in $\mathcal{F}(s) = A(s) \times \exp^{-j \times P(s)}$, where $A(s)$, $P(s)$ denote the amplitude and phase components, respectively. The Fourier transformation $\mathcal{F}(\cdot)$ and its inverse $\mathcal{F}^{-1}(\cdot)$  can be calculated with the FFT algorithm~\citep{elliott1982fast} effectively. Following the practice in~\citet{lv2022causality}: we intervene the amplitude by linearly interpolating between the amplitude of the original state $s$ and a state $s^\prime$ sampled randomly from a set which contains states where the non-causal factors have been removed (For Atari games, it is the displayed scored removed): 
\begin{equation}
    \hat{A}(s) = (1-\lambda)*A(s) + \lambda*A(s^\prime),
\end{equation}
where $\lambda \sim U(0, \epsilon)$ and $\epsilon$ adjusts the magnitude of intervention. Then we combine the perturbed amplitude with the original phase component to generate the intervened state $s^\textit{inter}$ by inverse Fourier transformation: $\mathcal{F}(s^\textit{inter}) = \hat{A}(s) \times \exp^{-j \times P(s)}, s^\textit{inter} = \mathcal{F}^{-1}(\mathcal{F}(s^\textit{inter})$.

\subsection{Clarifications: Causal Factors and Multi-task RL}

\subsubsection{Relationships between subsets of causal factors}

As is depicted in Fig.~\ref{fig:scm for causal explanations}, we expect causal factors to be as independent as possible. However, overlapping (between $\bar{\alpha}^i$ and $\bar{\alpha}^j$ subjecting to similar sub-tasks) is inevitable in many cases. For example, in the Monster-Treasure toy case that we study (see details in Appendix~\ref{sec:monstercase}), the agent receives a reward for reaching the treasure but incurs a penalty for landing on the monster; the agent (ego) becomes the overlapping part. Indeed, we use orthogonality and derive an objective function to encourage the subsets of causal factors ($\bar{\alpha}$) to be independent (if possible).

\subsubsection{Connection of RD to multi-task RL}

\textbf{Similarity}: 
\begin{itemize}
    \item Both RL with RD and multi-task RL contain the setting where there are multiple rewards (functions) from which corresponding policies can be learned.
\end{itemize}

\noindent \textbf{Dissimilarity}: 

\begin{itemize}
    \item RL with RD assumes the additivity of reward ($r = \sum_i r_i$), hence, we learn a global policy which is the summation of component policies (i.e., Q-function associated with each reward component). However, there is no such constraint toward the relation of reward functions designed for each task in the multi-task setting.
    \item In RL with RD setting, we learn all component policies in parallel, however, in multi-task RL, a single policy is generally sequentially updated across a sequence of tasks one by one.
\end{itemize}

Regarding the point of utilizing multiple reward channels to learn a multi-task RL agent, we believe it is a direction worth exploring. In RL with RD, the rollout is made by a global action which is derived from all component policies. However, when adapting multi-task RL to the setting of multiple reward channels, it may raise a further discussion about which rollout mechanism to employ (to collect trajectories) as it lacks a "global task" or "global policy" as we have in the RD setting. The intuitive way, for example, is to randomly choose the $i$-th task policy ($\pi(\cdot|s, z_i)$) to do the trajectory collection, and then use these trajectories to update other task policies, i.e., $\pi(\cdot|s, z_j), \, j \neq i$ with offline RL techniques.

\subsection{A Full List of Methods Used in Experiments}

\begin{table}[th]
\centering
\caption{The list of methods studied in experiments with varying learning features, encompassing aspects such as \textit{decomposing reward} (with full state or masked state factors), \textit{Q-agent learning} (with full state or masked state factors), \textit{knowledge of sub-reward values} in reward prediction (if applicable) and Q-learning, and \textit{the use of proposed desiderata} in factor learning. For example, the RD-pred method involves reward prediction and Q-agent learning with full state factors, and known sub-rewards, but it does not incorporate desiderata. RD, on the other hand, differs from RD-pred by not including reward prediction.
}
\scalebox{0.72}{
\begin{tabular}{l|>{\centering\arraybackslash}p{1.7cm} p{1.7cm}|>{\centering\arraybackslash}p{1.7cm} p{1.7cm}|>{\centering\arraybackslash}p{1.7cm}|>{\centering\arraybackslash}p{1.7cm}}
\toprule
\multirow{2}{*}{Method} & \multicolumn{2}{c|}{reward prediction $r^i$} & \multicolumn{2}{c|}{Q-value estimate $Q^i$}  & known sub-rewards  & desiderata losses\\
\cmidrule{2-7}
 & full state & sub-state & full state & sub-state & \\
\midrule
\midrule

RD &  \NA & \NA & \cmark & \xmark &\cmark & \xmark\\ 
RD-pred & \cmark & \xmark & \cmark & \xmark & \cmark & \xmark  \\
RD-pred-u & \cmark & \xmark & \cmark & \xmark & \xmark & \xmark \\
Q-Mask & \NA & \NA & \xmark & \cmark & \cmark & \cmark \\
Q-Mask Lite & \NA & \NA & \xmark & \cmark & \cmark & \xmark \\
R-Mask & \xmark & \cmark & \cmark & \xmark & \xmark & \cmark \\
R-Mask Lite & \xmark & \cmark & \cmark & \xmark & \xmark & \xmark \\

\bottomrule
\end{tabular}}

\label{table:all methods}
\end{table}

Table~\ref{table:all methods} lists all methods used in the experiments. In Q1, we compare RD with RD-pred to assess the impact of the auxiliary task of reward decomposition on the generation of explanation artefacts. Q2 involves a comparison between RD-pred-u and R-Mask, exploring the value of causal sufficiency of reward components. Q3 delves into the role of causal sufficiency concerning actions, comparing Q-Mask with R-Mask and RD-pred. Lastly, in Q4, we contrast R-Mask and Q-Mask with their Lite versions to elucidate the role of our proposed explanation criteria in learning disentangled, sparse causal factors.

\subsection{Deep Q-learning and Reward Decomposition}\label{sec:dqn-rd}

One of the fundamental approaches to learning the policy $\pi$ for an MDP involves initially acquiring knowledge about an action-value function \cite{Watkins:89}. This function encapsulates the anticipated cumulative discounted reward when the agent executes action $a_t$ within state $s_t$ and subsequently adheres to policy $\pi$ in the future. Formally, it can be expressed as
$Q(s_t, a_t) = \mathbb{E}_{\pi}[\, r_t + \gamma \max_{a_{t+1}}  Q(s_{t+1}, a_{t+1}) ]$,
where $\gamma$ denotes the discount factor. By determining the maximum value within the action-value function, an estimation of the optimal policy can be derived as $\hat{\pi}^* = \argmax_{a_t} Q(s_t, a_t)$.
Building upon the framework of deep Q-learning \cite{mnih2015humanlevel}, we approximate the value function $Q_\phi$ using a neural network-based function approximator that is parameterized by $\phi$. These parameters $\phi$ are iteratively refined by minimizing the loss function
\begin{equation*}
\begin{split}
    J(\phi) & = \mathbb{E}_{(s_t, a_t, r_t, s_{t+1}) \sim \mathcal{D}}[(r_t \, + \\
    & \gamma Q_{\phi^\prime} (s_{t+1}, \argmax_{a_{t+1}}Q_\phi(s_{t+1}, a_{t+1})) - Q_\phi(s_t, a_t))^2].
\end{split}
\end{equation*}
In this context, $Q_{\phi^\prime}$ denotes a target network, periodically synchronized with the main network $Q_\phi$ to stabilize learning \cite{vanHasselt15DeepReinforcement}.

When there are multiple reward components, we adopt a collection of $K \in \mathbb{N}$ Q-functions, each guided by an individual component $r^i$. The optimal (global) action $a^\ast_t$ corresponding to a state $s_t$ is identified as the one with the highest Q-value obtained by aggregating the Q-functions from all $K$ components $Q_{\phi^i}$, expressed as
$a^\ast_t = \argmax_{a_t^{\prime}} \sum_{i=1}^K Q_{\phi^{i}}(s_t, a_t^{\prime})$. 

\subsection{Evaluation Metrics for Explanations}\label{sec:eval-metrics}

\textbf{Fidelity.}  To assess the faithfulness of explanations objectively,
we calculate the \textit{fidelity} of the causal information transferred into the Q-agent, measured by the approximate information loss (see Sec.~\ref{sec:learning framework}) \small
$\mathcal{L}[ Q(a_t|s_t) \rightarrow Q(a_t|\bar{\alpha}^i_t) ] = \mathcal{H}[Q(a_t|s_t)|Q(a_t|\bar{\alpha}^i_t)]$
\normalsize
, i.e., the ability to make \textit{consistent} decisions when depending on the masked state (causal factor). The information loss (upper bound) can be measured as
\small
$\mathbb{E} \log p(a^*_t|\hat{a}^*_t) \le \log \mathbb{E} p(a^*_t|\hat{a}^*_t) \approx \log \frac{\#(a^*=\hat{a}^*)}{\#(a^*)}$
\normalsize
, which is the accuracy of directly estimating the full state decision
$a^*=\argmax_a \sum_i Q^i(a|s)$
with a distilled state $\hat{a}^* = \argmax_{a} \sum_i Q^i(a|\bar{\alpha}^i)$, computed by counting ($\#$) the consistency.

\noindent \textbf{Sparsity.} As the attention mask acts as an explanation artefact, it must be sufficiently obvious that users can appreciate it. Thus, sparse but distinct masks are preferred over dense ones (i.e., masks of value 1) for explanation purposes. For the evaluation of \textit{sparsity}, it involves a measure of information loss (the higher the better for sparsity) and information independence of sub-states.
The information loss can be approximately measured as the decrease of the information capacity (the lower the better) when the state is masked, i.e.,
\small
$\mathcal{L}(s \rightarrow \bar{\alpha}^i) \approx \mathcal{H}(\bar{\alpha}^i) \approx \mathbb{E} \frac{|\bar{\alpha}^i|}{|s|}$.
\normalsize

\noindent \textbf{Orthogonality.} For the benefit of interpretability, it is expected to obtain diverse attention masks each associated with a reward component, instead of all attention masks collapsing into a single mask. For the \textit{orthogonality} among states, we roughly evaluate their inter-independency as $I(\bar{\alpha}^i; \bar{\alpha}^j) = \mathcal{H}(\bar{\alpha}^i) + \mathcal{H}(\bar{\alpha}^j) - \mathcal{H}(\bar{\alpha}^i; \bar{\alpha}^j) \approx \frac{1}{|s|} \mathbb{E} (|\bar{\alpha}^i| + |\bar{\alpha}^j| - |\bar{\alpha}^i \cap \bar{\alpha}^j|)$, i.e., the overlap of masks.

\subsection{Monster-Treasure Toy-case}\label{sec:monstercase}

\begin{figure}[ht] 
\centering
\begin{minipage}{0.5\textwidth}
\centering
    \includegraphics[width=\textwidth]{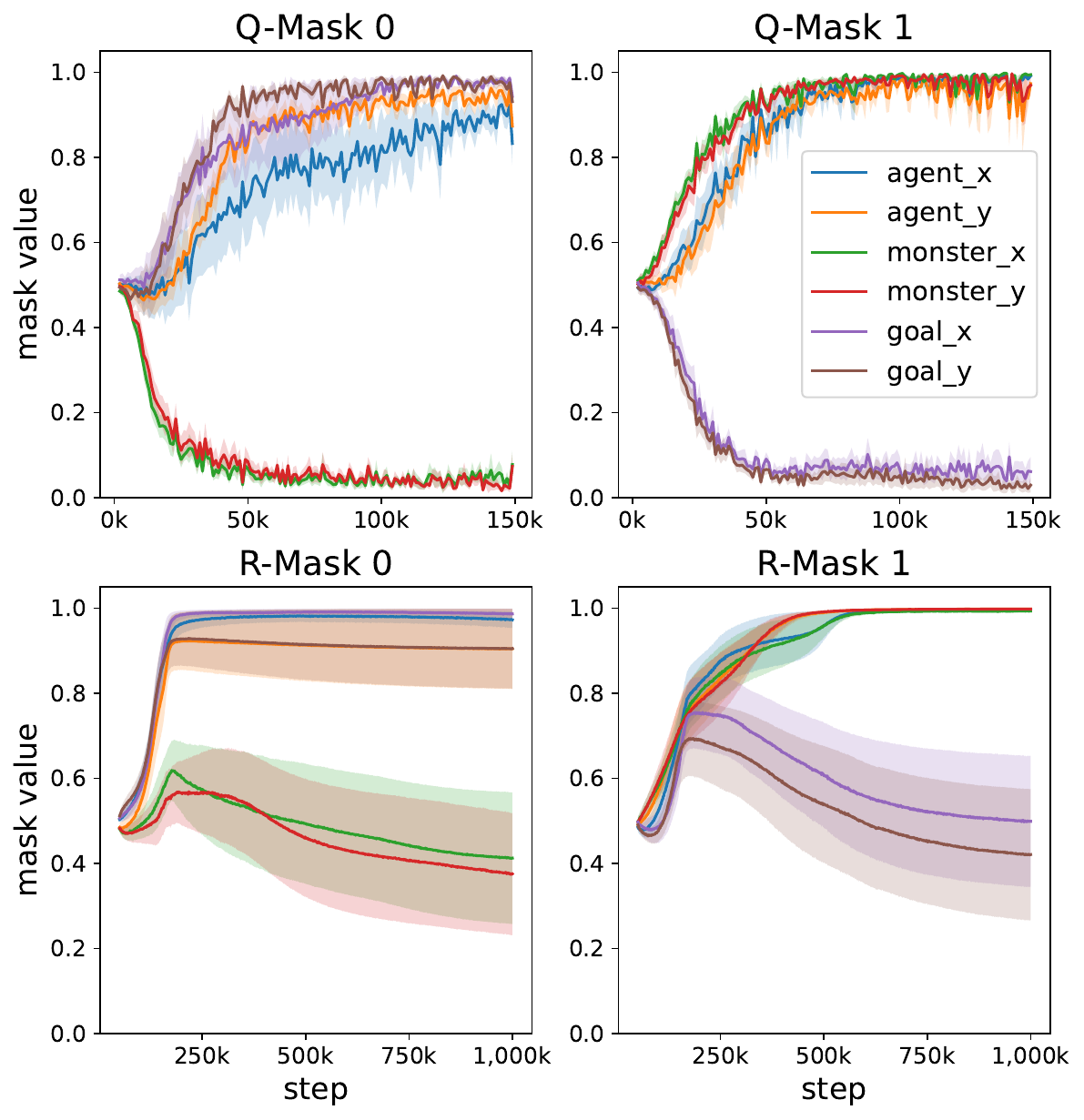}
    \captionof{figure}{Masks for the Monster-Treasure environment generated by Q-Mask and R-Mask. The plot shows the mean and standard error of ten runs. For R-Mask, the masks have been manually ordered so that mask 0 attends more to the treasure and mask 1 more to the monster.}
    \label{fig:mt_masks_full}
\end{minipage}%
\hfill
\begin{minipage}{0.48\textwidth}
    \centering
    \includegraphics[width=\textwidth]{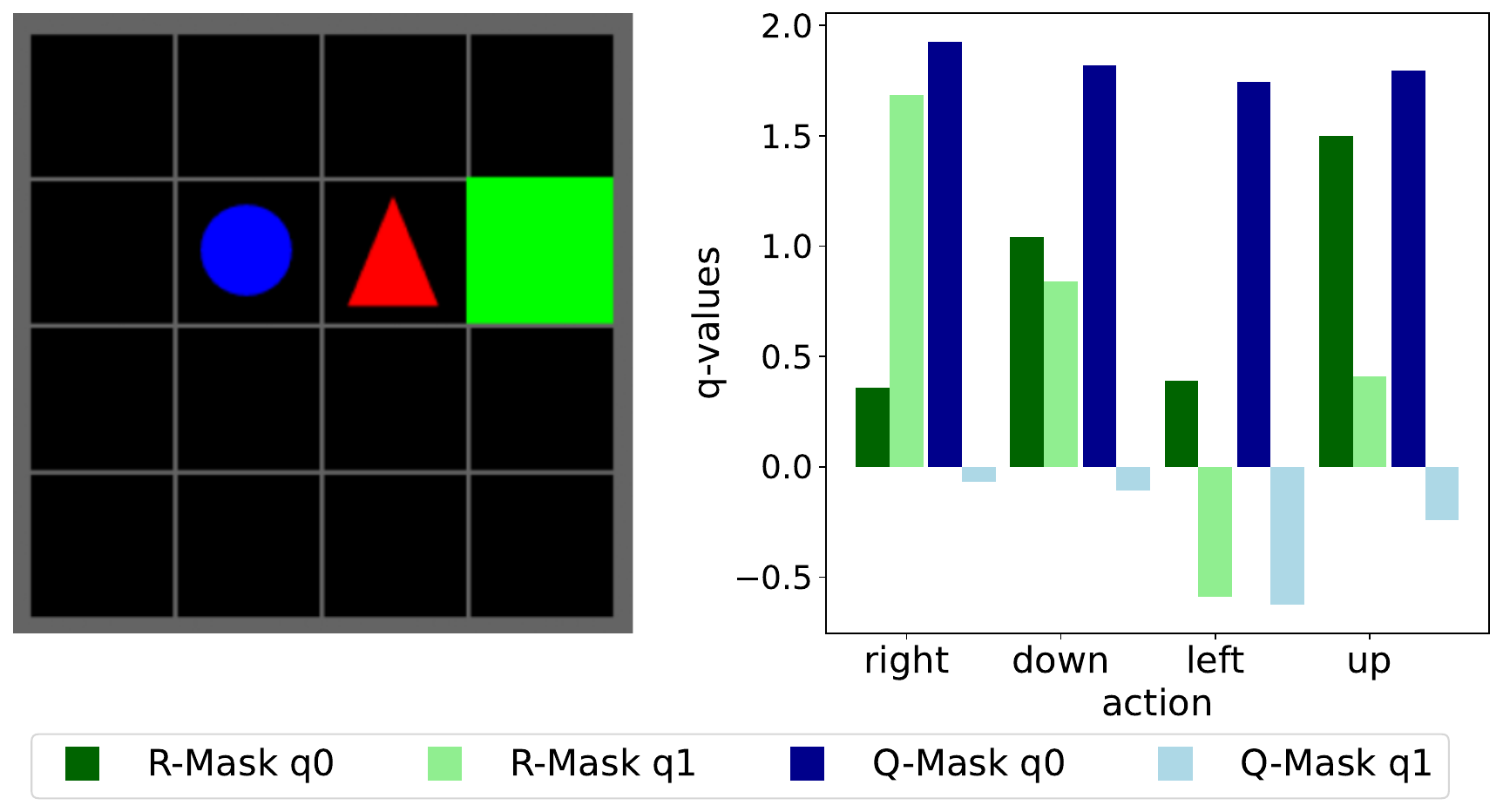}
    \caption{Example state for the Monster-Treasure environment with corresponding Q-values and reward predictions (Table~\ref{tab:reward prediction with R-Mask}). The agent (red arrow) is between the monster (blue circle) and the treasure (green square), and its choice is to move right. The component Q-values and component rewards add up to the full Q-values and the anticipated reward for each action.}
    \label{fig:mt_example_full}
\end{minipage}%

\end{figure}

This simple 2D mini-grid environment (Fig.~\ref{fig:mt_example_full}), initially introduced by \cite{gym_minigrid}, features a $4 \times 4$ grid hosting an agent with four possible movement directions, alongside a randomly spawned monster and treasure in each episode. The agent receives a reward $r^0 = 2$ for reaching the treasure's grid cell (goal) but incurs a $r^1 = -2$ penalty for landing on the monster's cell (i.e., $K=2$). The state includes the $x$- and $y$-coordinates of the agent, monster, and treasure, while the action space is going up, down, left and right.

To gain further insight into why R-Mask outperforms Q-Mask in generating high-quality masks (quantitatively and qualitatively) and determine whether this observation is coincidental, we evaluate them in a simplified scenario where we have complete access to ground truth causal factors for each sub-reward.

We depict the mask results learned by both Q-Mask and R-Mask methods in Fig.~\ref{fig:mt_masks_full}. It can be observed that mask values in Q-Mask gradually converge to optimal values, where the optimal monster mask is $\{1, 1, 1, 1, 0, 0\}$, i.e., estimated sub-state $s^{\text{monster}}$ = \{ $\text{agent}\_x$, $\text{agent}\_y$, $\text{monster}\_x$, $\text{monster}\_y$ \} under reward $r^1$, and the optimal treasure mask is $\{1, 1, 0, 0, 1, 1\}$ for $r^0$. However, R-Mask has difficulty distilling accurate sub-states, e.g., non-zero mask values for monster coordinates in the treasure mask 0.

In the depicted state (Fig.~\ref{fig:mt_example_full}), under Q-Mask, moving right yields the highest full Q-value (blue and light blue bars) for both rewards, while moving left (colliding with the monster) results in the lowest values. Each Q-agent in Q-Mask correctly focuses on its sub-state estimate when the agent chooses to move right toward the treasure cell. Instinctively, the agent's decision in that state is deemed trustworthy. In R-Mask, although imprecise masks are learned, when presenting the reward component estimates for a state under various actions in Table~\ref{tab:reward prediction with R-Mask} in Appendix~\ref{sec:ad-re-mt-env}, we observed the agent accurately estimating rewards. For instance, an estimate near 2 for a right move and close to -2 for a left move indicates trustworthy decision-making, favouring a right action.

\FloatBarrier



\subsection{Comparing R-Mask and Q-Mask with Their Lite Versions}\label{sec:com-rm-qm-lite}

R-Mask (Q-Mask) distinguishes itself from its lite version by incorporating explicit desiderata for exploring causal factors. The proposed indicators typically align with our perception of the generated explanations. Judging by the attention mask quality (e.g., Fig.~\ref{fig:gopher mask r-mask comparison 095}, Fig.~\ref{fig:gopher mask r-mask comparison2 095}, Fig.~\ref{fig:gopher mask rd2 comparison 015}, Fig.~\ref{fig:gopher mask q comparison} in Appendix~\ref{sec:add-re-go-env}), it becomes evident that R-Mask (Q-Mask) achieves a more favourable balance between these desiderata when contrasted with masks generated by their Lite versions, without the use of additional desiderata losses. 
For instance, in Fig.~\ref{fig:gopher mask q comparison}, the efficacy of Q-Mask's mask creation is evident: Q-Mask's Mask 0 highlights the agent's interaction with the gopher, while Mask 0 in Q-Mask Lite misses this. Similarly, Mask 1 in Q-Mask avoids irrelevant areas, such as the sky, unlike that in Q-Mask Lite which is less interpretable. Thus we cannot reliably trust them for explanation purposes. Another illustrative instance arises when comparing R-Mask Lite to R-Mask. Despite R-Mask Lite exhibiting superior fidelity scores compared to R-Mask, it generates masks that are dense and closely resembling one another (resulting in a high sparsity score of 0.932 and a substantial orthogonality score of 32.74). 

Masks created by R-Mask for Gopher environment exhibit a relatively high fidelity score\footnote{Achieving a fidelity score of $100\%$ can be readily demonstrated by setting masks to 1, yet it fails to be sparse.} and low sparsity score, indicating that ample but sparse information about the state is retained. This information proves predictive of both the agent's subsequent reward and its choice of action. However, in the MsPacman environment, R-Mask demonstrates lower fidelity. Given the intricate dynamics within MsPacman, including multiple moving characters (such as enemies) with which the agent must interact, as well as more reward sources ($K=3$), the process of rendering masks interpretable in MsPacman may encounter challenges.

\subsection{Case Studies}\label{sec:case-studies-appendix}

Two case studies are presented to demonstrate how diverse causal factors (attention masks) enhance our understanding of the agent's behaviour. We acknowledge that some conclusions are drawn from our subjective assessment of the generated explanations, and further refinement through a user study is a future consideration. Nonetheless, we leverage these case studies to illustrate how attention masks align with our expectations regarding the rationale behind the agent's actions.

For each scenario, we depict two examples of masks, juxtaposed for comparison. To understand the scenario the agent experienced and the masks correspond to, we overlay 4 consecutive (RGB) states by plotting each state with low transparency over one another. Thus, it is clear to see what each scenario represents.
The first scenario adheres to the critical state criterion, while the subsequent one illustrates the following state.

\textbf{R-Mask Attention Masks on Gopher.} 
We showcase attention masks learned by R-Mask in a critical scenario (Fig.~\ref{fig:important gopher masks r-mask}). To elaborate on why the agent prefers the ``LEFT'' move over the action ``LEFTFIRE'' at the scene, we first adopt reward difference explanation (RDX) as in~\cite{Juozapaitis19ExplainableReinforcement} to gain insight into the Q-value difference between the two actions under reward components gopher and ground, based on the bar plot of Q-values (rightmost in Fig.~\ref{fig:important gopher masks r-mask}; detailed computations in Appendix~\ref{sec:details-com-reward-de-ex}). RDX indicates moving left is preferable to the ``LEFTFIRE'' action due to a larger Q-value difference under the gopher reward component. This underscores the association between moving left and the presence of the gopher. Though it gives us the \textit{plain} reason, the diverse attention masks provided by R-Mask visually complement it and a broad look at Mask 0 (to the gopher and the agent) and Mask 1 (to the ground) gives us a visual intuition of what's going on. Mask 0 stays focused on the gopher and agent jointly, and as the agent nears the object, Mask 0 and Mask 1 follow and contract, as depicted in Fig.~\ref{fig:gopher masks2 with r=0.95}. 
This supports our hypothesis: the agent aims for double rewards through a sequence of actions: sprinting to the left before a ``UPFIRE'' action\footnote{As the gopher prepares to emerge from its hole, and the agent is above, executing a ``UPFIRE'' or ``FIRE'', creating a chance for a double reward.}. 

Notice the visual similarity between the two consecutive scenarios in Fig.~\ref{fig:important gopher masks r-mask}, with negligible pixel changes. Despite this, attention masks for each component adeptly capture and visually reflect subtle nuances,  which is essential for understanding the agent's one-step actions. This property holds for Q-Mask as well (see examples in Fig.~\ref{fig:important gopher masks q-mask}).

Beyond attentive masks, the R-Mask method accurately predicts reward components $r_i$ in the Gopher environment. For instance, in Fig. \ref{fig:gopher masks2 with r=0.95}, Mask 0 attends to the gopher and agent, predicting 0.827 (close to 0.8 actual value), while Mask 1 focuses on the ground, predicting 0.199 (close to 0.15 actual value). This reliability enables explaining the agent's preference for ``LEFTFIRE'' using R-Mask's attention masks.

\textbf{R-Mask Attention Masks on MsPacman.}
To further validate the ability of the proposed methods to mine the cause-effect relationships for more challenging environments when the reward causes are actually \textit{interdependent}, we test R-Mask on the MsPacman environment. Examining a critical scenario as depicted in Fig.~\ref{fig:important pacman masks r-mask}, Mask 0 significantly highlights Pacman and the blue ghost underneath, expanding as they converge. Hence, it visually reveals the rationale for the agent's downward movement choice.
Notably, in experiments, Mask 0 and Mask 1 often exhibit similarity, possibly due to the interplay between ``Ghost'' and ``EnergyPill'' rewards, where ``Ghost'' activation (i.e., is received) follows ``EnergyPill'' activation.
This inter-dependency between causal factors violates our assumption of additivity, making it challenging to decouple them from current learning objectives. However, the other causal components are still able to be extracted by the method.

Overall, we noticed a relatively low accuracy in predicting the reward for eating dots, possibly due to their significant magnitude difference (e.g., 0.25 vs. 5). Sparse and compact masks for this component were also rare, likely because of the dispersed dot distribution across the maze, making distinct masks less likely to appear (e.g., Mask 2 in Fig. \ref{fig:pacman masks r-mask at next state}).

\subsection{Adapting Causal Learning Across RL Domains: Text, Sound, and Tabular Data}

The adaptation process should be straightforward when dealing with reinforcement learning applications featuring multiple reward channels. It involves two key steps: 1. learning a causal representation (i.e., factors) of the raw state, separated from non-causal factors accessible through domain knowledge, and 2. developing maskers to selectively retrieve subsets of learned causal factors associated with each reward component.

As an illustration, consider applications involving auditory data. Initially, raw auditory data can be transformed into a spectrogram using a Short-Time Fourier Transform (STFT). Next, a neural network (NN) can be employed to extract a latent representation from it. Subsequently, our explanation approach can be applied.

\subsection{Details in the Implementation of Evaluation Metrics}
\subsubsection{The Choice of Critical State.} The selection of the critical state hinges on the criterion that the highest Q-value surpasses the second-highest Q-value by either $10\%$ or $15\%$.

\subsection{Details of Neural Network Architecture and Hyperparameters}\label{det-neu-net-arc}

\subsubsection{Training Flow.} 
\begin{figure}[thpb]
      \centering
      \includegraphics[scale=0.70]{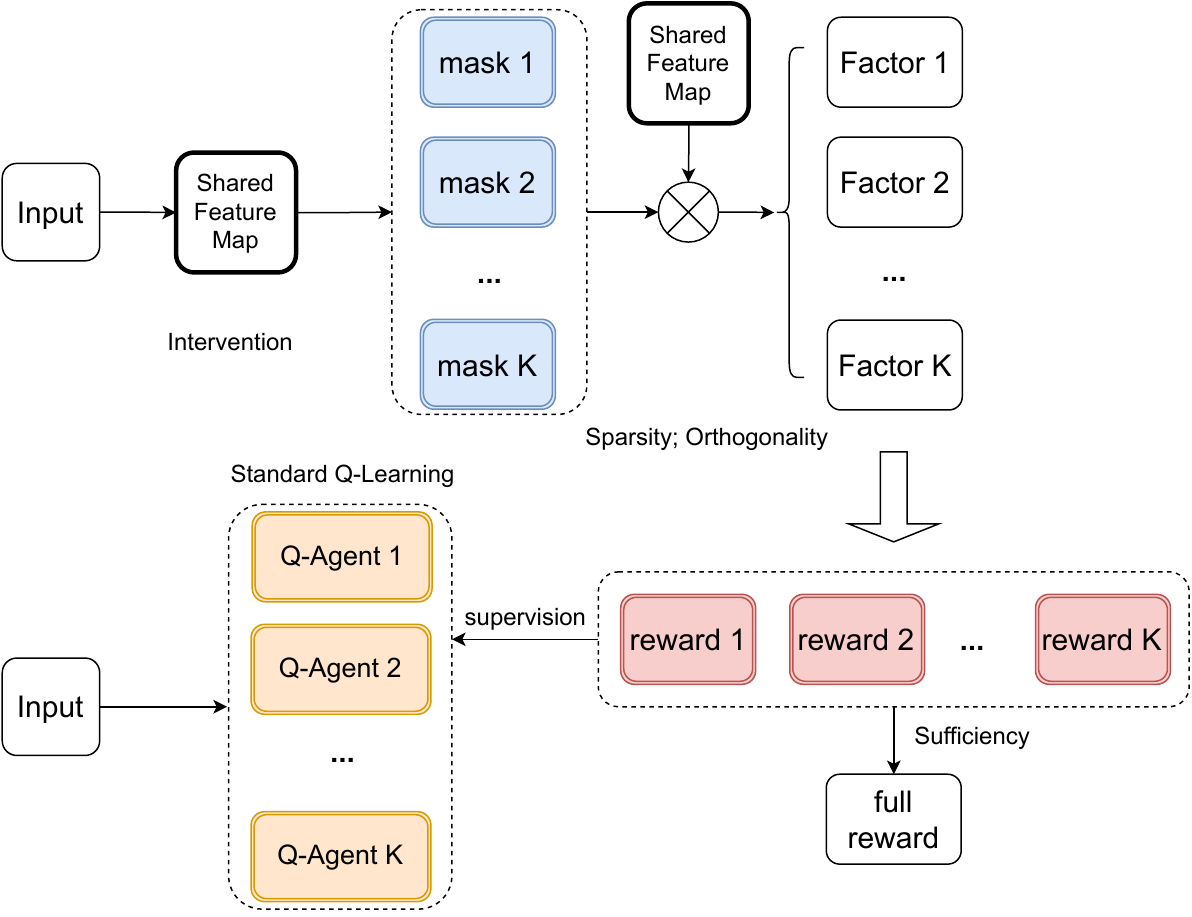}
      \caption{Training Flow in R-Mask: The neural modules (mask, Q-agent, and reward) to be learned are depicted by double-rounded rectangles, while the shared feature map (to be learned) upon which mask modules are constructed is represented by a bold-rounded rectangle. 
        Input is channelled through all $K$ mask modules, resulting in decomposed states. Subsequently, each reward module processes a decomposed state, generating a corresponding reward estimate. This yields a total of $K$ reward estimates, denoted as $r_{\theta^i}$. These estimates then serve as supervision signals, facilitating the update of each Q-function within the Q-agent module.
      }
      \label{fig:r-mask training flow}
\end{figure}

\begin{figure}[thpb]
      \centering
      \includegraphics[scale=0.71]{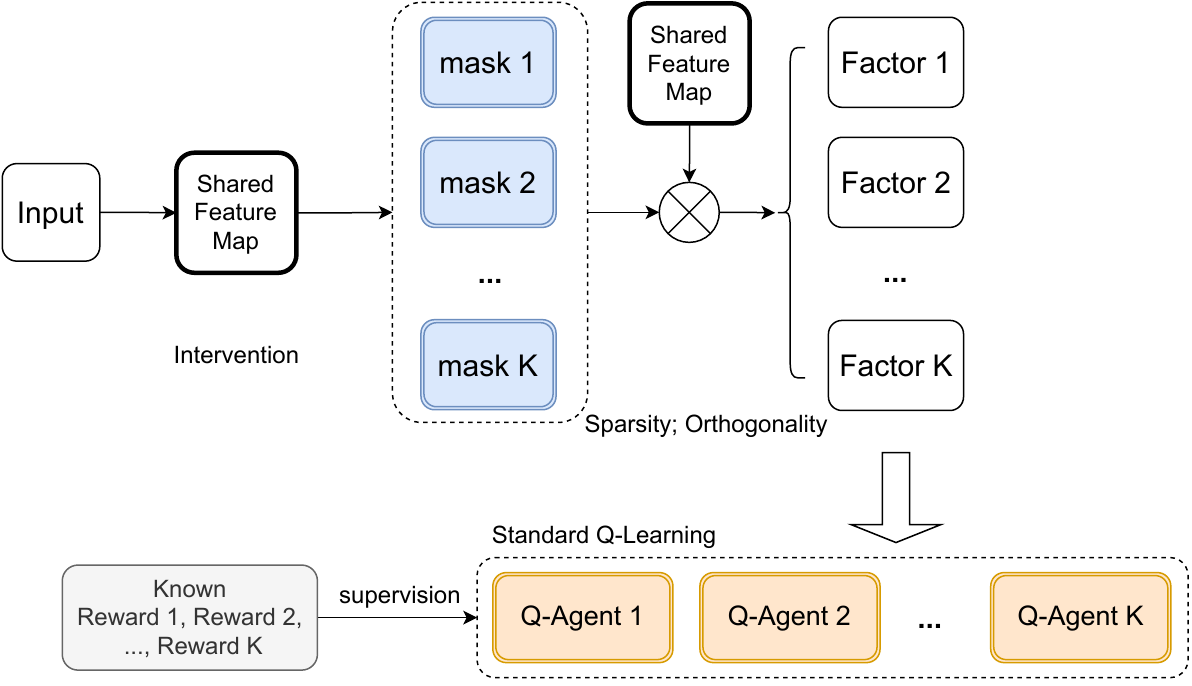}
      \caption{Training Flow in Q-Mask: The neural modules (mask and Q-agent) to be learned are depicted by double-rounded rectangles, while the shared feature map (to be learned) upon which mask modules are constructed is represented by a bold-rounded rectangle. 
The input is routed through all $K$ mask modules, generating decomposed states. Each Q-agent module then takes a decomposed state as input and is supervised by the corresponding ground truth reward component.
      }
      \label{fig:q-mask training flow}
\end{figure}

The training flow for R-Mask is illustrated in Fig.~\ref{fig:r-mask training flow}, while the training flow for Q-Mask is illustrated in Fig.~\ref{fig:q-mask training flow}.

\FloatBarrier

\subsubsection{Shared Feature Map.} 
\begin{figure}[!htb]
      \centering
      \includegraphics[scale=0.88]{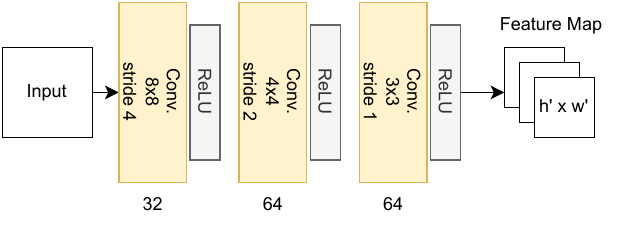}
      \caption{Conv-ReLU blocks in shared feature map (Conv: convolutional layer) in Fig.~\ref{fig:r-mask training flow}.
      }
      \label{fig:shared_feature_map}
\end{figure}

Both R-Mask and Q-Mask share a feature map structure depicted in Fig.~\ref{fig:shared_feature_map}. This structure comprises Conv-ReLU blocks with the following specifications: 1) Stride 4, $8\times8$ with 32 filters; 2) Stride 2, $4 \times 4$ with 64 filters; 3) Stride 1, $3\times3$ with 64 filters.

\subsubsection{Mask Module.} 
\begin{figure}[!htb]
      \centering
      \includegraphics[scale=0.88]{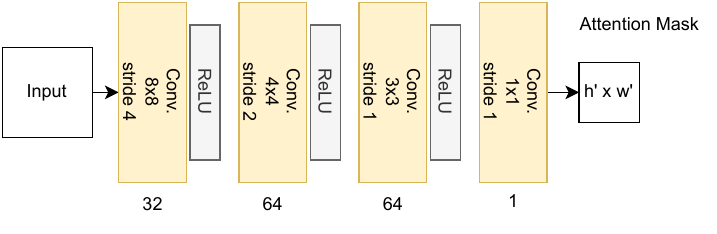}
      \caption{Conv-ReLU blocks in mask module (Conv: convolutional layer) in Fig.~\ref{fig:r-mask training flow}. A single $1\times1$ convolutional layer is employed to generate the attention mask output.
      }
      \label{fig:attention_mask}
\end{figure}

Each mask module follows a pattern as demonstrated in Fig.~\ref{fig:attention_mask}. This pattern encompasses Conv-ReLU blocks (the same as in Fig.~\ref{fig:shared_feature_map}) in conjunction with a $1\times1$ Conv layer, which produces the attention mask.


\subsubsection{Hyperparameters.} For Monster-Treasure and Atari environments, we choose to use Adam with a learning rate of $6.25e-5$ to update Q-functions, reward prediction networks and mask networks. Table \ref{tab:hyperparameters} lists the hyperparameters we use across all Atari games. The update frequencies $n_1, n_2, n_3, n_4$ are referred to in Algorithm \ref{alg:algorithm r-mask} and Algorithm \ref{alg:algorithm q-mask}, with the specific values being: $n_1 = 20$, $n_2 = 100$, $n_3 = 20$, and $n_4 = 20$. For the Monster-Treasure environment, we use $n_1 = 4$, $n_2 = 16$, $n_3 = 4$, and $n_4 = 4$. We run all experiments on a single GPU RTX 2080 Ti.

\begin{table}[!ht]
    \centering
    \caption{Preprocessing steps and hyperparameters}
    \begin{tabular}{l c}
        \toprule
        Parameter & Values\\
        \midrule
        \midrule
        Image Width & 84\\
        Image Height & 84\\
        GrayScaling & Yes\\
        Action Repetitions & 4\\
        Batch Size & 32\\
        Learning Rate & $6.25e-5$\\
        Discount Factor & 0.95\\
        \bottomrule
    \end{tabular}
    \label{tab:hyperparameters}
\end{table}

\subsection{Details in Computing Reward Decomposition Explanation (RDX)}\label{sec:details-com-reward-de-ex}

Section~\ref{sec:case studies} introduces RDX when explaining the agent's preference for the ``RIGHT'' move over the ``FIRE'' action in Fig.~\ref{fig:important gopher masks q-mask}. The computation of RDX is outlined as follows:

For any pair of actions, say $a_1$ and $a_2$, the difference in Q-values between the two actions under each component is represented as $\Delta_i(s, a_1, a_2) = Q_{\phi^{i}}(s, a_1) - Q_{\phi^{i}}(s, a_2)$. RDX serves as a quantitative measure, indicating the advantage or disadvantage of action $a_1$ compared to action $a_2$ under each component.

Considering Fig.~\ref{fig:important gopher masks r-mask}, we define $a_1$ as ``LEFT'' and $a_2$ as ``LEFTFIRE''. The Q-values are computed as follows: $Q(s, \text{LEFT}) = Q_\text{Gopher}(s, \text{LEFT}) + Q_\text{Ground}(s, \text{LEFT}) = 0.882 + 0.683$, and $Q(s, \text{LEFTFIRE}) = Q_\text{Gopher}(s, \text{LEFTFIRE}) + Q_\text{Ground}(s, \text{LEFTFIRE}) = 0.486 + 0.73$.

Under the Gopher reward component, we find \resizebox{0.5\textwidth}{!}{$\Delta_\text{Gopher} = Q_\text{Gopher}(s, \text{LEFT}) - Q_\text{Gopher}(s, \text{LEFTFIRE}) = 0.396$}. Under the Ground reward component, $\Delta_\text{Ground} = Q_\text{Ground}(s, \text{LEFT}) - Q_\text{Ground}(s, \text{LEFTFIRE}) = -0.047$. As $\Delta_\text{Gopher} \geq \Delta_\text{Ground}$, the agent's decision to move left rather than doing leftfire is influenced by the gopher, substantiating this behaviour.

\section{Pseudo-Code} \label{appendix:D}

Code is available at \url{https://github.com/LukasWill/causal-xrl}.

\subsection{Algorithm for R-Mask} \label{sec:r-mask-pseudo-code}
Algorithm~\ref{alg:algorithm r-mask} provides pseudo-code for R-mask on Atari environments which jointly learns component Q-functions and component rewards.

\begin{algorithm}[tb]
\caption{Reinforcement Learning with Masking (R-Mask)}
\label{alg:algorithm r-mask}


\KwIn{The number of reward components $K$, encoder parameters $\psi$, Q-function parameters $\phi^i$, parameters of reward prediction network $\theta^i$, parameters of mask network $\Psi^i$, and an empty replay buffer $\mathcal{D}$, where $i=1, 2, \cdots, K$. \\Set target parameters of Q-agent equal to main parameters $\phi_\text{target}^i \gets \phi^i $}

\BlankLine
\For{$t \leq \text{Total Steps}$}{
    Observe state $s_t$ and select action $a_t$ using $\epsilon$-greedy, $a_t = \argmax_{a_t^{\prime}} \sum_{i=1}^K Q_{\phi^{i}}(s_t, a_t^{\prime})$\;
    
    Execute $a_t$ in the environment\;
    
    Observe the next state $s_{t+1}$, reward $r_t$, and terminal signal $d$\;
    
    Store $(\,s_t, a_t, r_t, s_{t+1}, d ) \,$ in the replay buffer $\mathcal{D}$\;
    
    If $s_{t+1}$ is terminal, reset environment state\;
    
    \uIf{$t \geq \text{Learning Start Steps}$}{
        \uIf{$t \Mod{n_1} == 0$}{
            \tcp{Intervention, Sufficiency, Sparsity}
            
            Randomly sample batched transitions $B = \{ (s_t, a_t, r_t, s_{t+1}, d)\}$ from $\mathcal{D}$\;  
            
            Update parameters $\psi$ to maximize Eq.~\ref{eq:l0}, update parameters $\theta^i$ to minimize Eq.~\ref{eq:l1} and update parameters $\Psi^i$ to maximize Eq.~\ref{eq:l2} 
        }
        \uIf{$t \Mod{n_2} == 0$}{
            \tcp{Orthogonality}
            
            Randomly sample batched transitions $B = \{ (s_t, a_t, r_t, s_{t+1}, d)\}$ from $\mathcal{D}$\;

            Update parameters $\Psi^i$ to minimize Eq.~\ref{eq:l3} 
        }
        \uIf{$t \Mod{n_3} == 0$}{
            \tcp{Q-update}
            
            Randomly sample batched transitions $B = \{ (s_t, a_t, r_t, s_{t+1}, d)\}$ from $\mathcal{D}$\;
            
            Perform standard Q-learning using \underline{full reward} $r_t$ to update each parameter $\phi^i$ to minimize TD-error $\delta_1$\;
            \begin{equation*}
                \delta_1 = r_t + \gamma \sum_{i=1}^K Q_{\phi^{i}_\text{target}}(s_{t+1}, \argmax_{a^\prime} \sum_{i=1}^K Q_{\phi^{i}}(s_{t+1}, a^\prime)) - \sum_{i=1}^K Q_{\phi^{i}}(s_t, a_t)\;
            \end{equation*}
        }
        \uIf{$t \Mod{n_4} == 0$}{
            \tcp{Component Q-update}
            
            Randomly sample batched transitions $B = \{ (s_t, a_t, r_t, s_{t+1}, d)\}$ from $\mathcal{D}$\;
            
            Perform standard Q-learning using \underline{each estimate reward} $r_{\theta^i}$ to update each parameter $\phi^i$ to minimize TD-error $\delta_2$\;
            \begin{equation*}
            \delta_2 = r_{\theta^i} + \gamma Q_{\phi^{i}_\text{targ}}(s_{t+1}, a^\ast) - Q_{\phi^{i}}(s_t, a_t), \forall i.
            \end{equation*} 
            where $a^\ast = \argmax_{a^{\prime}} \sum_{i=1}^K Q_{\phi^{i}}(s_{t+1}, a^{\prime})$\;
        }
    }
}
\end{algorithm}

\subsection{Algorithm for Q-Mask} \label{sec:q-mask-pseudo-code}
Algorithm~\ref{alg:algorithm q-mask} provides pseudo-code for Q-mask on Atari environments which jointly learns component Q-functions and component rewards.

\begin{algorithm}[tb]
\caption{Reinforcement Learning with Masking (Q-Mask)}
\label{alg:algorithm q-mask}

\SetKwInput{Input}{Input}
\SetKwInput{KwData}{Data}
\SetKwInput{KwResult}{Result}
\SetKwInOut{Parameter}{Parameter}

\KwIn{The number of reward components $K$, encoder parameters $\psi$, Q-function parameters $\phi^i$, parameters of mask network $\Psi^i$, and an empty replay buffer $\mathcal{D}$, where $i=1, 2, \cdots, K$. \\Set target parameters of Q-agent equal to main parameters $\phi_\text{target}^i \gets \phi^i $}
\BlankLine

\For{$t \leq \text{Total Steps}$}{
  Observe state $s_t$ and select action $a_t$ using $\epsilon$-greedy: $a_t = \argmax_{a_t^{\prime}} \sum_{i=1}^K Q_{\phi^{i}}(s_t, a_t^{\prime})$\;
  
  Execute action $a_t$ in the environment\;
  
  Observe the next state $s_{t+1}$, rewards $\{ r^i_t \}$, and terminal signal $d$\;
  
  Store $(s_t, a_t, \{ r^i_t \}, s_{t+1}, d)$ in the replay buffer $\mathcal{D}$ \;
  
  If $s_{t+1}$ is terminal, reset the environment state\;

  \uIf{$t \geq \text{Learning Start Steps}$}{
    \uIf{$t \Mod{n_1} == 0$}{
    \tcp{Intervention, Sparsity}
        
      Randomly sample batched transitions $B = \{ (s_t, a_t, \{ r^i_t \}, s_{t+1}, d) \}$ from $\mathcal{D}$\;
      
      Update parameters $\psi$ to maximize Eq.~\ref{eq:l0} and update parameters $\Psi^i$ to maximize Eq.~\ref{eq:l2} 
    }

    \uIf{$t \Mod{n_2} == 0$}{
    \tcp{Orthogonality}
    
      Randomly sample batched transitions $B = \{ (s_t, a_t, \{ r^i_t \}, s_{t+1}, d) \}$ from $\mathcal{D}$ \;
      
      Update parameters $\Psi^i$ to minimize Eq.~\ref{eq:l3}\;
    }

    \uIf{$t \Mod{n_4} == 0$}{
    \tcp{Component Q-update}
    
      Randomly sample batched transitions $B = \{ (s_t, a_t, \{ r^i_t \}, s_{t+1}, d) \}$ from $\mathcal{D}$\;
      
      Perform standard Q-learning using \underline{ground truth sub-reward} $r_t^i$ to update each parameter $\phi^i$ and minimize TD-error $\delta$\;
      \begin{equation*}
          \delta = r_t^i + \gamma Q_{\phi^{i}_\text{target}}(s_{t+1}, a^\ast) - Q_{\phi^{i}}(s_t, a_t), \forall i
      \end{equation*}
        
      where $a^\ast = \argmax_{a^{\prime}} \sum_{i=1}^K Q_{\phi^{i}}(s_{t+1}, a^{\prime})$\;
    }
  }
}
\end{algorithm}

\end{document}